\documentclass{article}

\usepackage[utf8]{inputenc} \usepackage[T1]{fontenc}    \usepackage{hyperref}       \usepackage{url}            \usepackage{booktabs}       \usepackage{amsfonts}       \usepackage{nicefrac}       \usepackage{color}
\usepackage{epsfig}
\usepackage{amsthm}
\usepackage{amsmath}
\usepackage{float}
\usepackage{graphicx}
\usepackage{wrapfig}
\usepackage{makecell}
\usepackage{caption}
\usepackage{subcaption}
\usepackage{bm}
\usepackage{multirow}
\usepackage{microtype}
\usepackage{mathtools}
\usepackage{tikz}
\usepackage{gensymb}
\usepackage{bbm}
\usepackage{chngcntr}
\usepackage{amssymb}
\usepackage{enumitem}
\usepackage{placeins}

\usepackage{algorithm}
\usepackage{algorithmic}
\usepackage{xcolor} 
\usepackage[T1]{fontenc} 
\usepackage{listings}
\usepackage{xspace}
\usepackage{setspace}

\usepackage{hyperref}

\usepackage[accepted]{icml2023}

\def\tape{interface\xspace}
\def\Tape{Interface\xspace}
\def\modelname{RINs\xspace}
\def\fullname{Recurrent Interface Networks\xspace}

\newcommand{\latents}{Z}
\newcommand{\X}{X}
\newcommand{\by}{{\mkern-2mu\times\mkern-2mu}}

\icmltitlerunning{\fullname}

\begin{document}

\twocolumn[
\icmltitle{Scalable Adaptive Computation for Iterative Generation}

\icmlsetsymbol{equal}{*}
\icmlsetsymbol{workdoneatgoo}{\dag}

\begin{icmlauthorlist}
\icmlauthor{Allan Jabri}{goo,berk,workdoneatgoo}
\icmlauthor{David J.\ Fleet}{goo}
\icmlauthor{Ting Chen}{goo}

\end{icmlauthorlist}

\icmlaffiliation{goo}{Google Brain, Toronto.}
\icmlaffiliation{berk}{Department of EECS, UC Berkeley}

\icmlcorrespondingauthor{Ting Chen}{iamtingchen@google.com}
\icmlcorrespondingauthor{Allan Jabri}{ajabri@berkeley.edu}

\icmlkeywords{Neural Interface Networks, Iterative Generation}

\vskip 0.3in
]

\printAffiliationsAndNotice{Code: github.com/google-research/pix2seq \textsuperscript{\dag}Work done as a student researcher at Google.}

\begin{abstract}
    Natural data is redundant yet predominant architectures tile computation uniformly across their input and output space. We propose the \textit{Recurrent Interface Networks} (RINs), an attention-based architecture that decouples its core computation from the dimensionality of the data, enabling adaptive computation for more scalable generation of high-dimensional data.
    \modelname focus the bulk of computation (i.e. global self-attention) on a set of \textit{latent} tokens, using cross-attention to read and write (i.e. \textit{route}) information  between latent and data tokens. 
    Stacking RIN blocks allows bottom-up (data to latent) and top-down (latent to data) feedback, leading to deeper and more expressive routing.
While this routing introduces challenges, this is less problematic in recurrent computation settings where the task (and routing problem) changes gradually, such as iterative generation with diffusion models.
    We show how to leverage recurrence by conditioning the latent tokens at each forward pass of the reverse diffusion process with those from prior computation, i.e. latent self-conditioning.
    \modelname yield state-of-the-art pixel diffusion models for image and video generation, scaling to $1024\by1024$ images without cascades or guidance,
    while being domain-agnostic and up to 10$\times$ more efficient than 2D and 3D U-Nets.
\end{abstract} \section{Introduction}

\begin{figure}[t]
\vspace{-1mm}
\begin{center}
\includegraphics[width=0.99\linewidth]{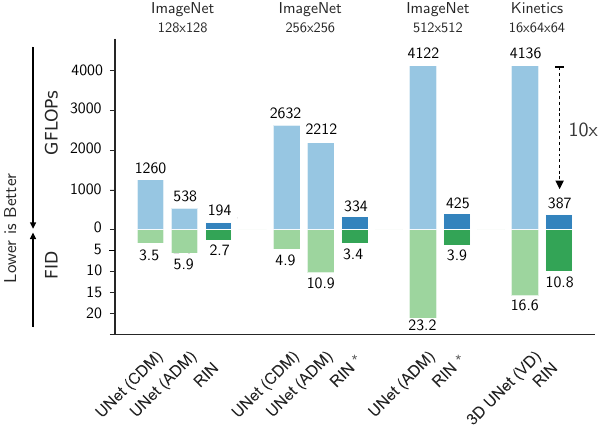} 
\end{center}
\vspace{-3mm}
\caption{\label{fig:key_result}\modelname outperform U-Nets widely used in state-of-the-art image and video diffusion models, while being more efficient and domain-agnostic. Our models are simple pixel-level denoising diffusion models without cascades as in (CDM \cite{ho2021cascaded}) or guidance (as in
ADM \cite{dhariwal2021diffusion} and VD \cite{ho-videodiffusion}). $*$: uses input scaling \cite{chen2023noise}.
}
\vspace{-6mm}
\end{figure}

\begin{figure*}[th!]
    \begin{center}
\includegraphics[width=0.93\linewidth]{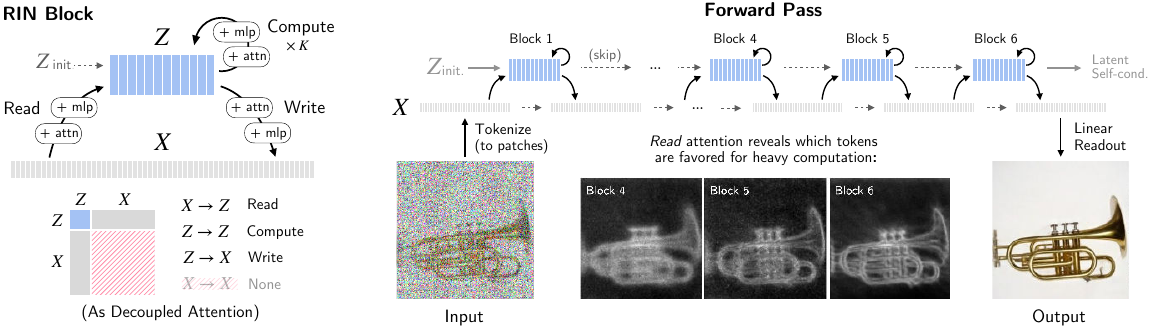} 
\end{center}
    \vspace{-4mm}
    \caption{\label{fig:arch_overview} \textbf{Overview of \fullname.} The input is tokenized to form the \tape $X$. A stack of blocks route information between $X$ and latents $Z$, avoiding quadratic pairwise interactions between tokens in X (bottom left). Note that $|Z| < |X|$, and most computation is applied to $Z$, which allows for scaling to large X. The network's read attention maps reveals how tokens are favored for latent computation (right), when trained for a task like diffusion generative modeling. See Figure~\ref{fig:attn_vis} for more visualizations. 
}
    \vspace{-4mm}
\end{figure*}

The design of effective neural network architectures has been crucial
to the success of deep learning~\cite{krizhevsky2012imagenet,he2016deep,vaswani2017attention}.
Influenced by modern accelerator hardware, predominant architectures,
such as convolutional neural networks ~\cite{fukushima1988neocognitron, lecun1989backpropagation,he2016deep} and Transformers~\cite{vaswani2017attention}, allocate computation 
in a fixed, uniform manner over the input data (e.g., over image pixels, image patches, or token sequences).
Information in natural data is often distributed unevenly, or exhibits redundancy, so 
it is important to ask how to allocate computation in an adaptive manner to improve scalability.
While prior work has explored more dynamic and input-decoupled computation, e.g., networks with auxiliary memory ~\cite{dai2019transformer,rae2019compressive} and global units ~\cite{zaheer2020big,burtsev2020memory,jaegle2021perceiver,jaegle2021perceiverio},
general architectures that leverage adaptive computation to effectively scale to tasks with large input and output spaces remain elusive.

Here, we consider this
issue 
as it manifests in high-dimensional generative modeling tasks, such as image and video generation.
When generating an image with a simple background, an adaptive architecture should ideally be able to allocate
computation to regions with
complex objects and textures, rather than regions with little or no structure (e.g., the sky).
When generating video, one should exploit temporal redundancy, allocating less computation to static regions.
While such non-uniform computation becomes more crucial in higher-dimensional data, achieving it
efficiently is challenging, especially with modern hardware that favours fixed computation graphs and dense matrix multiplication.

To address this challenge, we propose an architecture, dubbed {\fullname} (\modelname).
In {\modelname} (Fig.~\ref{fig:arch_overview}), hidden units are partitioned into the \textit{\tape} $X$ and \textit{latents} $Z$.
\Tape units are locally connected to the input and grow linearly with input size.
In contrast, latents are decoupled from the input space, forming a more compact representation on which the bulk of computation operates.
The forward pass proceeds as a stack of blocks that read, compute, and write:
in each block, information is routed from \tape tokens (with cross-attention) into the latents for high-capacity global processing (with self-attention), and updates are written back to \tape tokens (with cross-attention). 
Alternating computation between latents and \tape allows for processing at local and global levels, accumulating context for better routing.
As such, {\modelname} allocate computation more dynamically
than uniform models, scaling better when 
information is unevenly distributed across the input and output, as is common in natural data.

This decoupling introduces additional challenges, which can overshadow benefits if the latents are initialized without context in each forward pass, leading to shallow and less expressive routing.
We show this can be mitigated in scenarios involving recurrent computation, where the task and routing problem change gradually such that persistent context can be leveraged across iterations to
in effect form a deeper network. 
In particular, we consider iterative generation of images and video with denoising diffusion models~\cite{sohl2015deep,ho2020denoising,song2020denoising,song2021scorebased}.
To leverage recurrence, we propose latent self-conditioning as a ``warm-start'' mechanism for latents: 
instead of reinitializing latents 
at each forward pass, we use latents from previous iterations as additional context, like a recurrent network but without requiring backpropagation through time.

Our experiments with diffusion models show that \modelname outperform 
U-Net architectures for image and video generation, as shown in Figure~\ref{fig:key_result}. 
For class-conditional ImageNet models, 
from $64\by 64$ up to $1024\by 1024$, 
\modelname outperform leading diffusion models that use cascades~\cite{ho2021cascaded} or  guidance~\cite{dhariwal2021diffusion,ho2021classifierfree}, while
consuming up to $10\by$ fewer FLOPs per inference step.
For video prediction, \modelname surpass leading approaches~\cite{ho-videodiffusion} on the Kinetics600 benchmark while reducing the FLOPs of each step by $10\by$.

Our contributions are summarized as follows:
\begin{itemize}[topsep=-2pt, partopsep=0pt, leftmargin=12pt, parsep=0pt, itemsep=3pt]
    \item We propose \modelname, a domain-agnostic architecture capable of adaptive computation for scalable generation of high dimensional data.
    \item We identify recurrent computation settings in which \modelname thrive and advocate latent self-conditioning to amortize
    the challenge of routing.
    \item Despite reduced inductive bias, this leads to performance and efficiency gains over U-Net diffusion models for image and video generation.
\end{itemize}

\section{Method}

\begin{figure*}[t!]
  \vspace{0mm}
  \begin{center}
\includegraphics[width=0.999\linewidth]{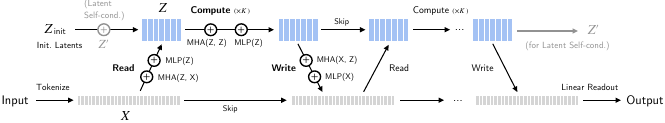}
  \end{center}
  \vspace{-5mm}
  \caption{\label{fig:arch_detail}
  \textbf{The RIN computation graph.} \modelname stack blocks that read, compute, and write. 
  \textit{Read} operations load information into latents with cross-attention.
  \textit{Compute} operations exchange information across latent tokens with self-attention and across channels with token-wise MLPs.
  \textit{Write} operations update the \tape with information from the latents with cross-attention, and mix information across channels with token-wise MLPs.
  Latent self-conditioning (gray lines) allows for propagation of latent context between iterations.
  }
  \vspace{-4mm}
\end{figure*}

In \modelname, the \tape is locally connected to the input space and initialized via a form of tokenization (e.g., patch embeddings), while the latents are decoupled from data and initialized as learnable embeddings. The basic RIN block allocates computation by \textit{routing} information between the \tape and the latents.
By stacking multiple blocks, we can update the \tape and latents repeatedly, such that bottom-up and top-down context can inform routing in the next block (see Fig. \ref{fig:arch_detail}). 
A linear readout function predicts the network's output from the final \tape representation.

Since the \tape is tied to data, it grows linearly with input size and may be large (e.g., thousands of vectors), while the number of latent units can be much smaller (e.g., hundreds of vectors).
The computation operating directly on the \tape (e.g. tokenization, read, write) is uniform across the input space, but is designed to be relatively light-weight, for minimal uniform computation. 
The high-capacity processing is reserved for the latents, formed by reading information from the \tape selectively, such that most computation can be adapted to the structure and content of the input.

Compared to convolutional nets such as U-Nets~\cite{ronneberger2015u}, \modelname do not rely on fixed downsampling or upsampling for global computation. 
Compared to Transformers~\cite{vaswani2017attention}, \modelname operate on sets of tokens with positional encoding for similar flexibility across input domains, but avoid pairwise attention across tokens to reduce compute and memory requirements per token.
Compared to other decoupled architectures such as PerceiverIO~\cite{jaegle2021perceiver,jaegle2021perceiverio}, alternating computation between \tape and latents enables more expressive routing without a prohibitively large set of latents.

While \modelname are versatile, their advantages are more pronounced in recurrent settings, where inputs may change gradually over time such that it is possible to propagate persistent context to further prime the routing of information. 
Therefore, here we focus on the application of \modelname to iterative generation with diffusion models.

\subsection{Background: Iterative Generation with Diffusion} 

We first provide a brief overview of diffusion models~\citep{sohl2015deep,ho2020denoising,song2020denoising,song2021scorebased,kingma2021variational,chen2022analog}.
Diffusion models learn a series of state transitions to map noise $\bm \epsilon$ from a known prior distribution to $\bm x_0$ from the data distribution.
To learn this (reverse) transition from noise to data, a forward transition from $\bm x_0$ to $\bm x_t$ is first defined:
\vspace{-0.5mm}
\begin{equation*}
\bm x_t = \sqrt{\gamma(t)} ~\bm x_0 + \sqrt{1-\gamma(t)} ~\bm \epsilon,
\end{equation*}
where $\bm \epsilon \sim \mathcal{N}(\bm 0, \bm I)$, $t\sim \mathcal{U}(0, 1)$, and $\gamma(t)$ is a monotonically decreasing function from 1 to 0.
Instead of directly learning a neural net to model the transition from $\bm x_t$ to $\bm x_{t-\Delta}$, one can learn a neural net $f(\bm x_t, t)$ to predict $\bm \epsilon$ from $\bm x_t$, and then estimate $\bm x_{t-\Delta}$ from the estimated $\tilde{\bm \epsilon}$ and  $\bm x_{t}$.
The objective for $f(\bm x_t, t)$ is thus the $\ell_2$ regression loss:
\begin{equation*}
\mathbb{E}_{t\sim\mathcal{U}(0, 1), \bm \epsilon\sim\mathcal{N}(\bm 0, \bm 1)}\|f(\sqrt{\gamma(t)} ~\bm x_0 + \sqrt{1-\gamma(t)} ~\bm \epsilon, t) - \bm \epsilon \|^2.
\end{equation*}
To generate samples from a learned model, we follow a series of (reverse) state transition $\bm x_1\rightarrow \bm x_{1-\Delta} \rightarrow  \cdots  \rightarrow \bm x_0$.
This is done by iteratively applying the denoising function $f$ on each state $\bm x_t$ to estimate $\bm \epsilon$, and hence $\bm x_{t-\Delta}$, using transition rules as 
 in DDPM~\citep{ho2020denoising} or DDIM~\citep{song2020denoising}.
As we will see, the gradual refinement of $\bm x$ through repeated application of the denoising function is a natural fit for \modelname.
The network takes as input a noisy image $\bm x_t$, a time step $t$, and an optional conditioning variable e.g. a class label $y$, and then outputs the estimated noise $\tilde{\bm \epsilon}$.

\subsection{Elements of \fullname}

We next describe the major components of \modelname (Fig.~\ref{fig:arch_detail}).

\textbf{\Tape Initialization.}
The \tape is initialized from an input $\bm x$, such as an image $\bm x_\text{image} \in \mathbb{R}^{h\times w \times 3}$, or video $\bm x_\text{video} \in \mathbb{R}^{h\times w\times l \times 3}$ by tokenizing $\bm x$ into a set of $n$ vectors $X\in\mathbb{R}^{n\times d}$. For example, we use a linear patch embedding similar to~\cite{dosovitskiy2020image} to convert an image into a set of patch tokens; for video, we use 3-D patches. To indicate their location, patch embeddings are summed with (learnable) positional encodings. Beyond tokenization, the model is domain-agnostic, as $X$ is simply a set of vectors.

\textbf{Latent Initialization.}
The latents $Z \in \mathbb{R}^{m\times d'}$ are (for now) initialized as 
learned embeddings,
independent of the input.
Conditioning variables, such as class labels and time step $t$ of diffusion models, are mapped to embeddings; in our experiments, we simply concatenate them to the set of latents, since they only account for two tokens.

\begin{figure*}[t!]
  \begin{center}
\includegraphics[width=0.925\linewidth]{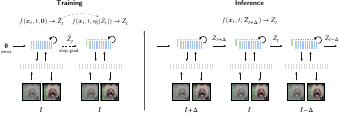} 
  \end{center}
  \vspace{-3mm}
  \caption{\label{fig:latent_sc} 
  \textbf{Latent Self-Conditioning for Diffusion Models with \modelname.}
  (Left) During training, latents for self-conditioning are first estimated with a forward pass of the denoising network (with zeros as previous latents);
  we then condition the denoising network with these estimated latents by treating them as latents of the previous iteration 
  (without back-propagating through the estimated latents). 
  (Right) During sampling, we start with zero latents, and use computed latents at each time-step to condition the next time-step.
  }
  \vspace{-4mm}
\end{figure*}

\textbf{Block.}
The RIN block routes information between $X$ and $Z$ with components of Transformers:
\vspace{2mm}
\begin{align*}
    \textrm{\textbf{Read:}} \mspace{20mu}
    \latents &= Z + \textrm{MHA}(\latents, \X)\\ 
    \latents &= Z + \textrm{MLP}(\latents) \\ \noalign{\vskip3pt}
    \textrm{\textbf{Compute}:} \mspace{20mu}
    \latents &= Z + \textrm{MHA}(\latents, \latents) \\
    (\times K) \mspace{30mu} 
    \latents &= Z + \textrm{MLP}(\latents) \\ \noalign{\vskip3pt}
    \textrm{\textbf{Write:}} \mspace{20mu}
    \X &= X + \textrm{MHA}(\X, \latents)\\
    \X &= X + \textrm{MLP}(\X)
\end{align*}
\vspace{-1mm}

MLP denotes a multi-layer perceptron, and MHA(Q, K) denotes multi-head attention with queries Q, and keys and values K.\footnote{See~\cite{vaswani2017attention} for details about multi-head attention, which extends single-head attention defined as $\textrm{Attention}(\latents, \X) = \textrm{softmax}(\latents W_Q W_K^\top \X^\top ) \X W_V$. $\textrm{MLP}(\latents) = \sigma (\latents W_1 + b_1) W_2 + b_2$ where $\sigma$ is the GELU activation function \cite{hendrycks2016gaussian}. $W$ are learned linear projections. We apply LayerNorm~\cite{ba2016layer} on the queries of MHA operations.} 

The depth of each block $K$ controls the ratio of computation occurring on the \tape and latents.
From the perspective of information exchange among hidden units, MHA propagates information across vectors (i.e. between latents, or between latents and \tape), while the MLP (applied vector-wise, with shared weights) mixes information across their channels.
Note that here computation on the \tape is folded into the write operation, as MHA followed by an MLP.

RIN blocks can be stacked to allow latents to accumulate context and write incremental updates to the \tape. To produce output predictions, we apply a readout layer (e.g. a linear projection) to the corresponding \tape tokens to predict local outputs (such as patches of images or videos). The local outputs are then combined to form the desired output (e.g., patches are simply reshaped into an image). A detailed implementation is given in Appendix~\ref{sec:net_implementation} (Alg~\ref{alg:net_implementation}).

\subsection{Latent Self-Conditioning}

\modelname rely on routing information to dynamically allocate compute to parts of the input. Effective routing relies on latents that are specific to the input, and input-specific latents are built by reading \tape information. This iterative process can incur additional cost that may overshadow the benefits of adaptive computation, especially if the network begins without context, i.e. from a ``cold-start''.
Intuitively, as humans, we face a similar ``cold-start'' problem under changes in the environment, requiring gradual familiarization of new state to enhance our ability to infer relevant information.
If contexts switch rapidly without sufficient time for ``warm-up'', we repeatedly face the cost of adapting to context.
The ``warm-up'' cost in \modelname can be similarly amortized in recurrent computation settings 
where inputs gradually change while global context persists. We posit that in such settings, there exists useful context in the latents accumulated in each forward pass.

\textbf{Warm-starting Latents.} With this in mind, we propose to ``warm-start'' the latents using latents computed at a previous step.
The initial latents at current time step $t$ are the sum of the learnable embeddings $Z_{init}$ (independent of the input), and a transformation of previous latents computed in the previous iteration $t'$:
$$\latents_{t} = \latents_{init} + \textrm{LayerNorm}(\latents_{t'} + \textrm{MLP}(\latents_{t'})) ~, $$
where $\textrm{LayerNorm}$ is initialized with zero scaling and bias, so that $\latents_{t} = \latents_{init}$ early in training.

In principle, this relies on the existence of latents from a previous time step, $Z_{t'}$, and requires unrolling iterations and learning with backpropagation through time, which can hamper scalability.
A key advantage of diffusion models is that the chain of transitions decomposes into conditionally independent steps allowing for highly parallelizable training, an effect we would like to preserve.
To this end, we draw inspiration from the self-conditioning technique of~\cite{chen2022analog}, which conditions a denoising network at time $t$ with its own unconditional prediction for time $t$.

Concretely, consider the conditional denoising network $f(\bm x_t, t, Z_{t'})$ that takes as input $\bm x_t$ and $t$, as well as context latents $Z_{t'}$. During training, with some probability, we use $f(\bm x_t, t, \bm 0)$ to directly compute the prediction $\bm \tilde{\epsilon}_t$. 
Otherwise, we first apply $f(\bm x_t, t, \bm 0)$ to obtain latents $\tilde{Z}_{t}$ as an estimate of $Z_{t'}$, and compute the prediction with $f(\bm x_t, t, \text{sg}(\tilde{Z}_{t}))$.
Here, $\text{sg}$ is the stop-gradient operation, used to avoid back-propagating through the latent estimates. At inference time, we directly use latents from previous time step $t'$ to initialize the latents at current time step $t$, i.e., $f(\bm x_t, t, {Z}_{t'})$, in a recurrent fashion. This bootstrapping procedure marginally increases the training time ( < 25\% in practice, due to the stop-gradient), but has a negligible cost at inference time.
In contrast to self-conditioning at the data level~\cite{chen2022analog}, here we condition on the latent activations of the neural network,
so we call it 
\textit{latent self-conditioning}.

Figure~\ref{fig:latent_sc} illustrates the training and sampling process with the proposed latent self-conditioning. 
Algorithms~\ref{alg:train} and~\ref{alg:sample} give the proposed modifications to training and sampling of the standard diffusion process.
Details of common functions used in the algorithms can be found in Appendix~\ref{sec:alg_details}.

\begin{algorithm}[tb]
\caption{\small Training \modelname with Latent Self-Cond.
}
\label{alg:train}
\definecolor{codeblue}{rgb}{0.25,0.5,0.5}
\definecolor{codekw}{rgb}{0.85, 0.18, 0.50}
\lstset{
  backgroundcolor=\color{white},
  basicstyle=\fontsize{7.2pt}{7.2pt}\ttfamily\selectfont,
  columns=fullflexible,
  breaklines=true,
  captionpos=b,
  commentstyle=\fontsize{7.2pt}{7.2pt}\color{codeblue},
  keywordstyle=\fontsize{7.2pt}{7.2pt}\color{codekw},
  escapechar={|}, 
}
\vspace{-1mm}
\begin{lstlisting}[language=python]
def train_loss(x, self_cond_rate, latent_shape):
  # Add noise.
  t = uniform(0, 1)
  eps = normal(mean=0, std=1)
  x_t = sqrt(gamma(t)) * x + sqrt(1-gamma(t)) * eps

  # Compute latent self-cond estimate.
  |\color{blue}latents = zeros(latent\_shape)|
  |\color{blue}if uniform(0, 1) < self\_cond\_rate:|
    |\color{blue}\_, latents = rin((x\_t, latents), t)|
    |\color{blue}latents = stop\_gradient(latents)|
    
  # Predict and compute loss.
  eps_pred, _ = |\color{blue}rin|((x_t, |\color{blue}latents|), t)
  loss = (eps_pred - eps)**2
  return loss.mean()
\end{lstlisting}
\vspace{-2mm}
\end{algorithm}
 \section{Experiments}

We demonstrate that \modelname improve state-of-the-art performance on benchmarks for image generation and video prediction with pixel-space diffusion models. In all experiments, we do not use guidance. 
For each benchmark, we also compare the number of floating point operations (GFLOPs) across methods; as we will see,
\modelname are also more efficient. Samples and further visualizations are provided in Appendix~\ref{sec:samples} and the supplementary material.

\subsection{Implementation Details} 

\textbf{Noise Schedule.}
Similar to~\cite{kingma2021variational, chen2022analog}, we use a continuous-time noise schedule function $\gamma(t)$. By default we use a cosine schedule, as in previous work~\cite{nichol2021improved} but find it is sometimes unstable for higher resolution images.
We therefore explore schedules 
based the sigmoid function with different temperature, which shifts weight away from the tails of the noise schedule. We use a default temperature of 0.9, and its effect is ablated in our experiments. Detailed implementation of noise schedules and ablations are provided in Appendix~\ref{sec:alg_details}. For larger images, we also report models trained using input scaling\cite{chen2023noise,chen2022generalist}.

\begin{algorithm}[t]
\caption{\small Sampling with Latent Self-Cond.}
\label{alg:sample}
\definecolor{codeblue}{rgb}{0.25,0.5,0.5}
\definecolor{codekw}{rgb}{0.85, 0.18, 0.50}
\lstset{
  backgroundcolor=\color{white},
  basicstyle=\fontsize{7.2pt}{7.2pt}\ttfamily\selectfont,
  columns=fullflexible,
  breaklines=true,
  captionpos=b,
  commentstyle=\fontsize{7.2pt}{7.2pt}\color{codeblue},
  keywordstyle=\fontsize{7.2pt}{7.2pt}\color{codekw},
  escapechar={|}, 
}
\vspace{-1mm}
\begin{lstlisting}[language=python]
def generate(steps):
  x_t = normal(mean=0, std=1)
  |\color{blue}latents = zeros(latent\_shape)|
  
  for step in range(steps):
    # Get time for current and next states.
    t = 1 - step / steps
    t_m1 = max(1 - (step + 1) / steps, 0)
        
    # Predict eps.
    eps_pred, |\color{blue}latents| = |\color{blue}rin|((x_t, |\color{blue}latents|), t)
        
    # Estimate x at t_m1.
    x_t = ddim_or_ddpm_step(x_t, eps_pred, t, t_m1)
    
  return x_t
\end{lstlisting}
\vspace{-2mm}
\end{algorithm}

\begin{figure}[h!]
\vspace{-2mm}
\begin{center}
\hspace{-4mm}
\subfloat[$\gamma(t)$]{
\label{fig:noise_schedule}
\includegraphics[width=0.45\linewidth]{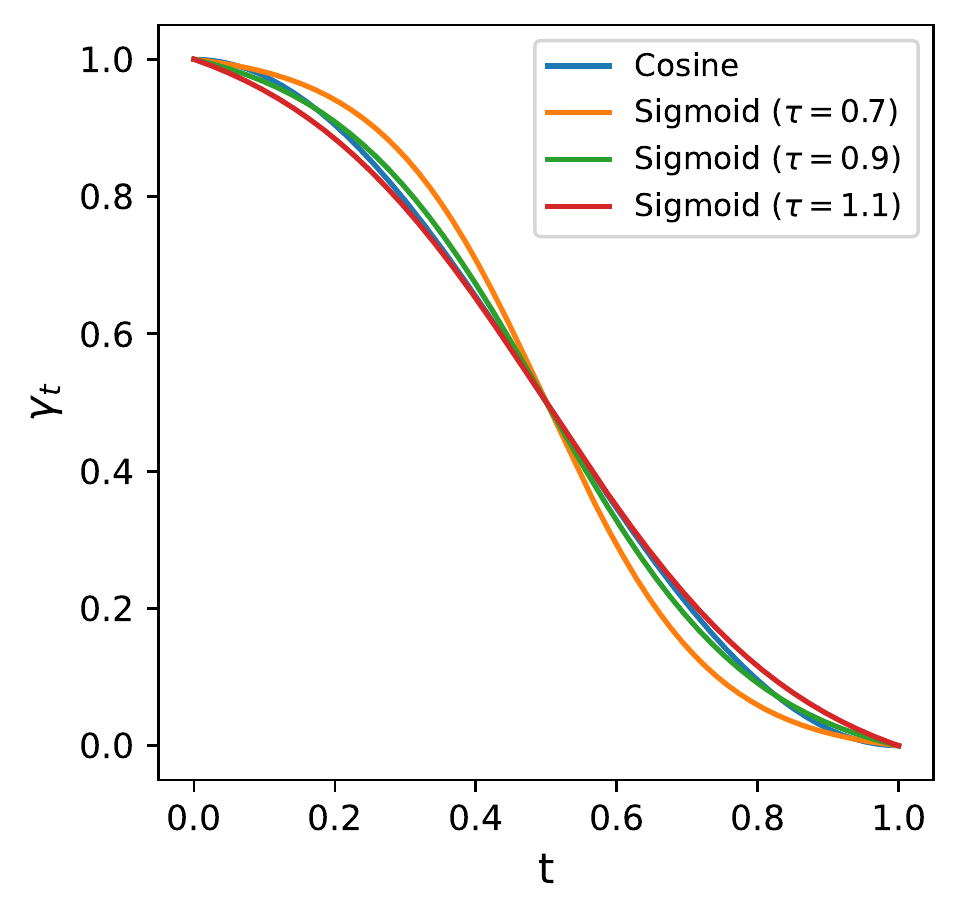} 
}
\subfloat[$\log(\gamma(t)/(1-\gamma(t)))$]{
\label{fig:patchsize_ablation}
\includegraphics[width=0.45\linewidth]{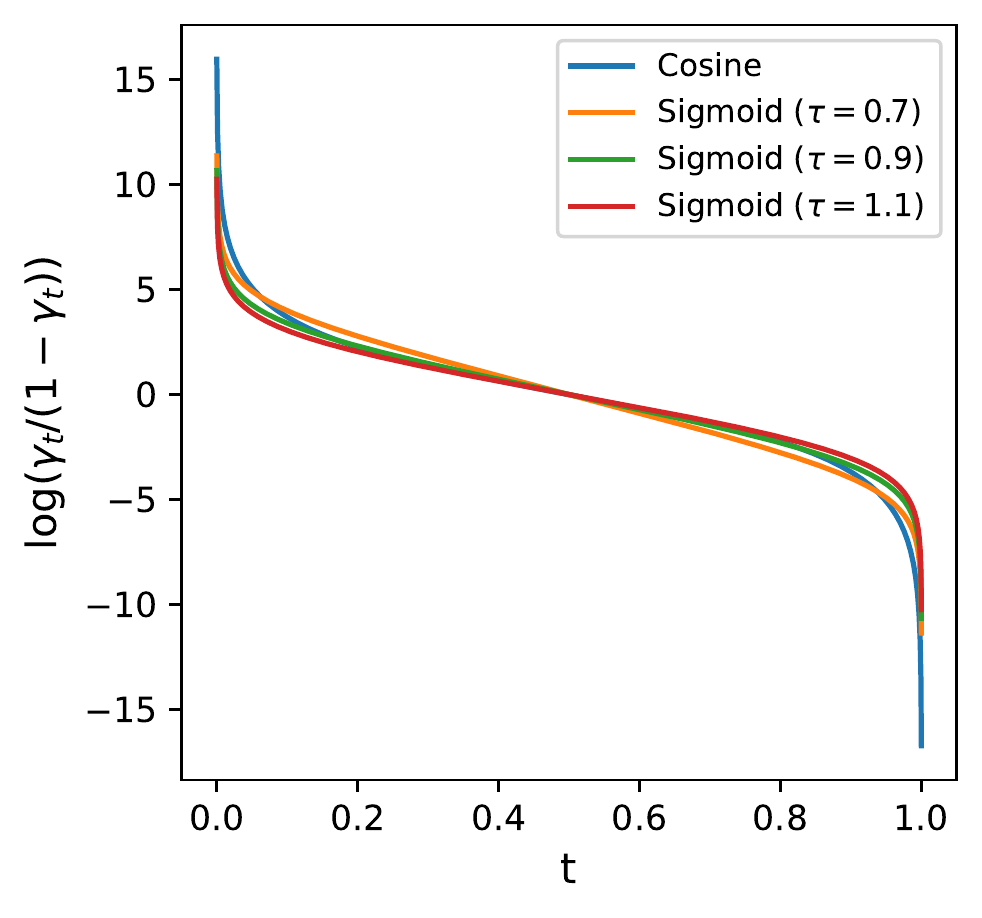}
}
\end{center}
\vspace{-4mm}
\caption{Compared to the cosine schedule, sigmoid (with appropriate $\tau$) can place less weight on noise levels on the tails.}
\vspace{-2mm}
\end{figure}

\begin{figure*}[t!]
\begin{center}
\includegraphics[width=0.95\linewidth]{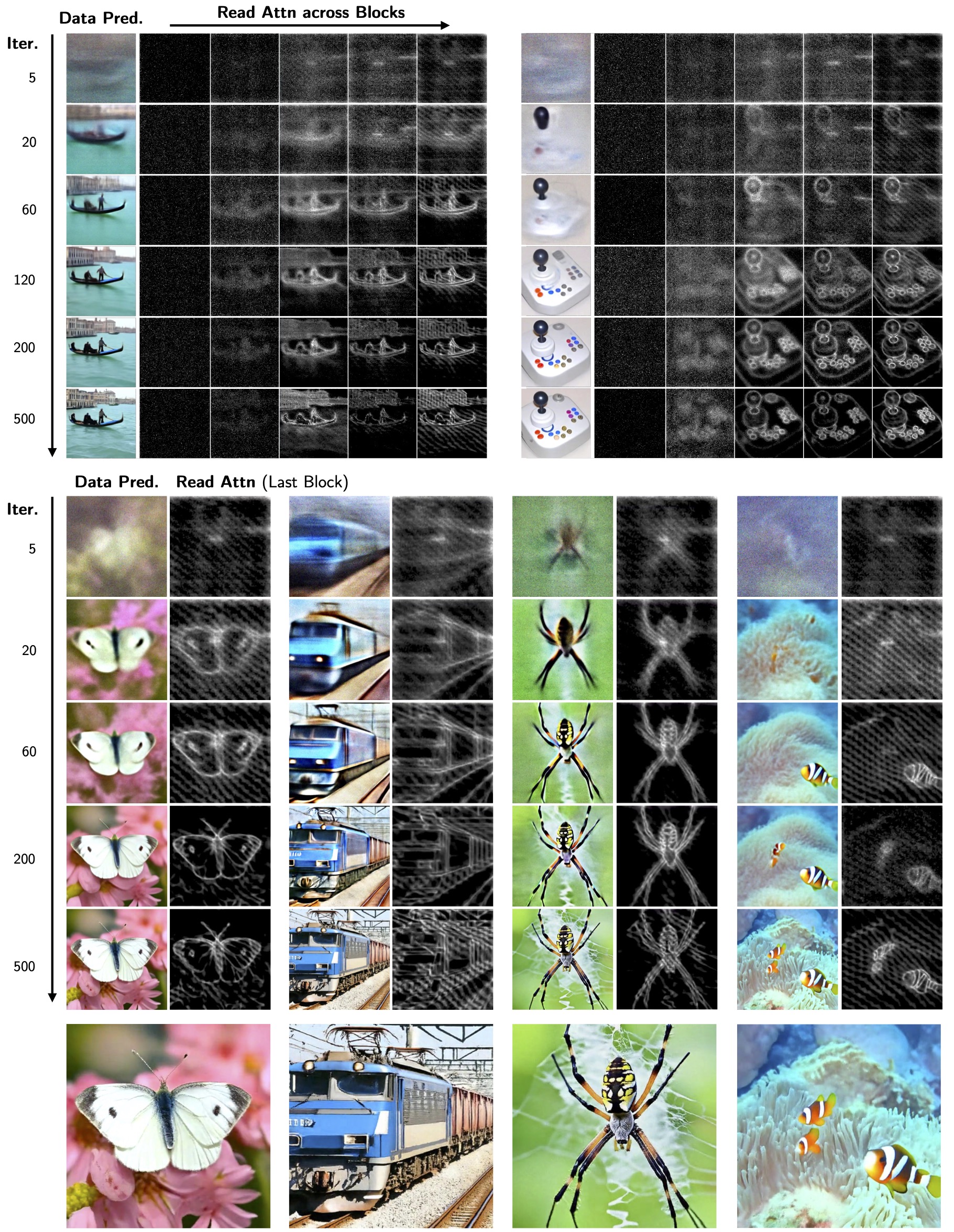}
\end{center}
\vspace{-3mm}
\caption{\label{fig:attn_vis} \textbf{Visualizing Adaptive Computation.} The read attention reveals which information is routed into latents for heavy computation. We visualize the read attention (averaged across latents) at each block (top) or the last block (bottom), at selected steps of the reverse process when generating ImageNet $512\by512$ samples. While it is similar across samples in early iterations, it becomes more sparse and data-specific, focusing latent computation on more complex regions.}
\vspace{-3mm}
\end{figure*}

\textbf{Tokenization and Readout.} For image generation, we tokenize images by extracting non-overlapping patches followed by a linear projection. We use a patch size of 4 for $64\by 64$ and $128\by 128$ images, and 8 for larger images. To produce the output, we apply a linear projection to \tape tokens and unfold each projected token to obtain predicted patches, which we reshape to form an image. 

For video, we tokenize and produce predictions in the same manner as images; for $16\by 64 \by 64$ inputs, we use {2$\by$4$\by$4} patches, resulting in $2048$ tokens. For conditional generation, during training, the context frames are provided as part of the input, without noise added.
During sampling, the context frames are held fixed. 

\begin{table}[t]
\vspace{-3mm}
\caption{RIN configurations for each task.}
\label{tab:rin_config}
\vspace{1mm}
\begin{center}
\begin{small}
\begin{tabular}{ccccccr}
\toprule
 & $128$px & $256$px & $512$px & $1024$px & Kinetics \\
\midrule
$|Z|$ & $128$ & $256$  & $256$ & $256$ & $256$ \\
dim($Z$) & $1024$ & $1024$  & $768$  & $768$  & $1024$\\
$|X|$ & $1024$ & $1024$  & $4096$ & $16384$  & $2048$\\
dim($X$) & $512$ & $512$  & $512$ & $512$ & $512$ \\
Blocks & 6 & 6 & 6 & 6 & 6\\
Depth $K$ & 4 & 4 & 6 & 8 & 4\\
Tokens & $4\by4$ & $8\by 8$ & $8\by 8$ & $8\by 8$ & $2\by 4 \by 4$ \\
\bottomrule
\end{tabular}
\end{small}
\end{center}
\vspace{-4mm}
\end{table}

Table~\ref{tab:rin_config} compares model configuration across tasks; note the ratio of $|Z|$ and $|X|$. See Appendix~\ref{sec:train_details} for model and training hyper-parameters, and Appendix~\ref{sec:net_implementation} for pseudo-code.

\subsection{Experimental Setup}

For image generation, we mainly use the ImageNet dataset~\cite{russakovsky2015imagenet}. For data augmentation, we only use center crops and random left-right flipping. We also use CIFAR-10~\cite{cifar10} to show the model can be trained with small datasets.
For evaluation, we follow common practice, using FID~\cite{heusel2017gans} and Inception Score~\cite{salimans2016improved} as metrics computed on 50K samples, generated with 1000 steps of DDPM.

For video prediction, we use the Kinetics-600 dataset~\cite{carreira2018short} at 16$\times$64$\times$64 resolution. For evaluation, we follow common practice~\cite{ho-videodiffusion} and use FVD~\cite{unterthiner2018towards} and Inception Scores  computed on 50K samples, with 400 or 1000 steps of DDPM.

\subsection{Comparison to SOTA}

\begin{table}[t]
\vspace{-2mm}
\caption{Comparison to leading approaches for Class-Conditional Generation on ImageNet. $\dagger$: use of class guidance,  1: \cite{dhariwal2021diffusion}, 2: \cite{ho2021classifierfree}, 3: \cite{ho2021cascaded}.
}
\label{tab:image_sota}
\vspace{1mm}
\begin{center}
\begin{small}
\begin{tabular}{lcccc}
\toprule
Method & FID $\downarrow$ & IS $\uparrow$ & GFLOPs & Param(M) \\
\midrule

{IN $64\by 64$}  &&&& \\
\cmidrule{1-1}
ADM $^1$  & -- &  2.07 &210& 297 \\
CF-guidance $^{2\dagger}$  & 1.55 & 66.0 &  -- & --\\
CDM $^3$  & 1.48 & 66.0 &  -- & --\\
{RIN}  & \textbf{1.23} & \textbf{66.5} &106& 281 \\
\midrule

IN $128\by 128$ &&&& \\
\cmidrule{1-1}
ADM $^1$  & 5.91 & -- &538& 386 \\
ADM + guid. $^{1\dagger}$  & 2.97 & -- & >538&  >386\\
CF-guidance $^{2\dagger}$  & \textbf{2.43} & \textbf{156.0} &  --  & -- \\
CDM $^3$  & 3.51 & 128.0 &1268& 1058  \\
{RIN}  & 2.75 & 144.1 &194&  410 \\
\midrule

IN $256\by 256$ &&&& \\
\cmidrule{1-1}
ADM $^1$  & 10.94 & 100.9 &2212&  553 \\
ADM + guid.$^{1\dagger}$  & 4.59 & 186.7 &>2212& >553\\
CDM$^3$  & 4.88 & 158.7 &2620& 1953\\
{RIN}  & {4.51} & {161.0} &334& 410 \\
{RIN + inp. scale}  & \textbf{3.42} & \textbf{182.0} & 334 & 410 \\

\midrule
IN $512\by 512$ &&&& \\
\cmidrule{1-1}
ADM $^1$  & 23.2 & 58.1 & 4122&  559 \\
ADM + guid.$^{1\dagger}$  & 7.72 & 172.7 & >4122 & >559\\
{RIN + inp. scale}  & \textbf{3.95} & \textbf{216.0} & 415 & 320 \\

\midrule
IN $1024\by 1024$ &&&& \\
\cmidrule{1-1}
{RIN + inp. scale}  & \textbf{8.72} & \textbf{163.9} & 1120 & 412 \\

\bottomrule
\end{tabular}
\end{small}
\end{center}
\vspace{-6mm}
\end{table}

\begin{figure*}[!t]
     \centering
    \begin{subfigure}[b]{0.39\textwidth}
     \vspace{1mm}
     \includegraphics[width=\textwidth]{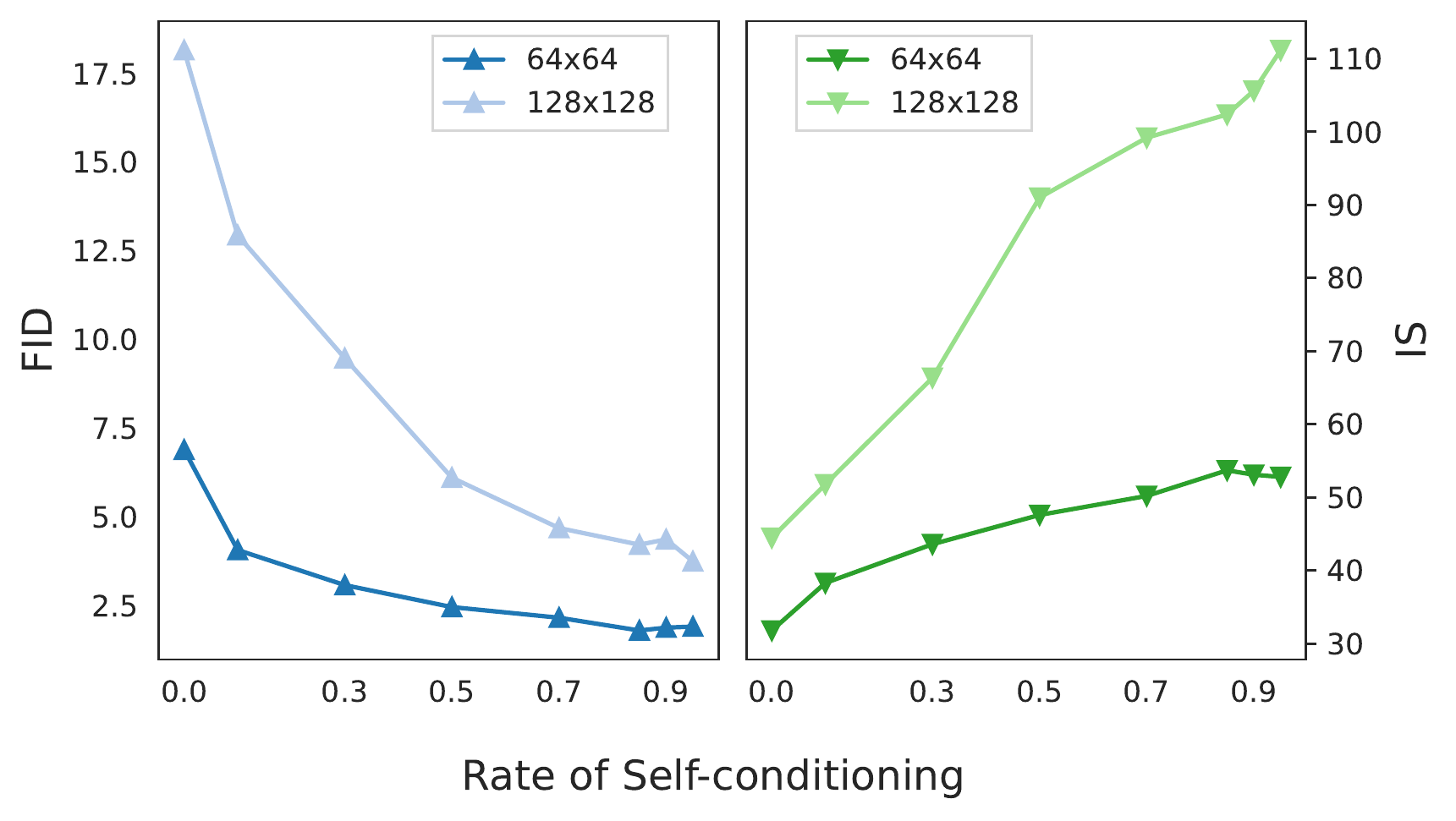}
     \caption{\label{fig:sc_rate}Latent self-condition}
    \end{subfigure}
    \begin{subfigure}[b]{0.37\linewidth}
     \vspace{-0.0mm}
     \includegraphics[width=\textwidth]{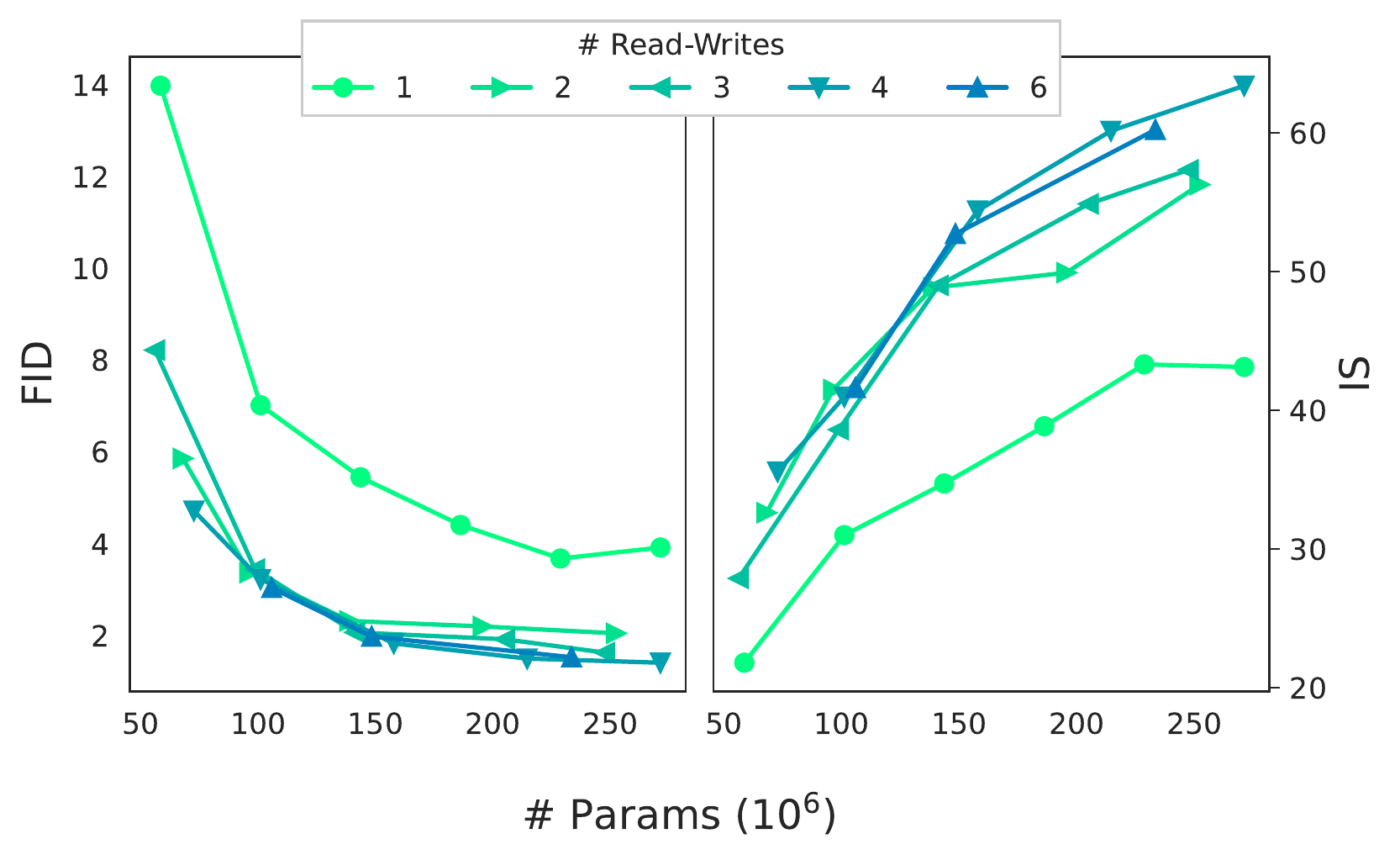}
\caption{\label{fig:tradeoff_read_write}Number of Blocks}
    \end{subfigure}
    \begin{subfigure}[b]{0.225\linewidth}
     \includegraphics[width=\textwidth]{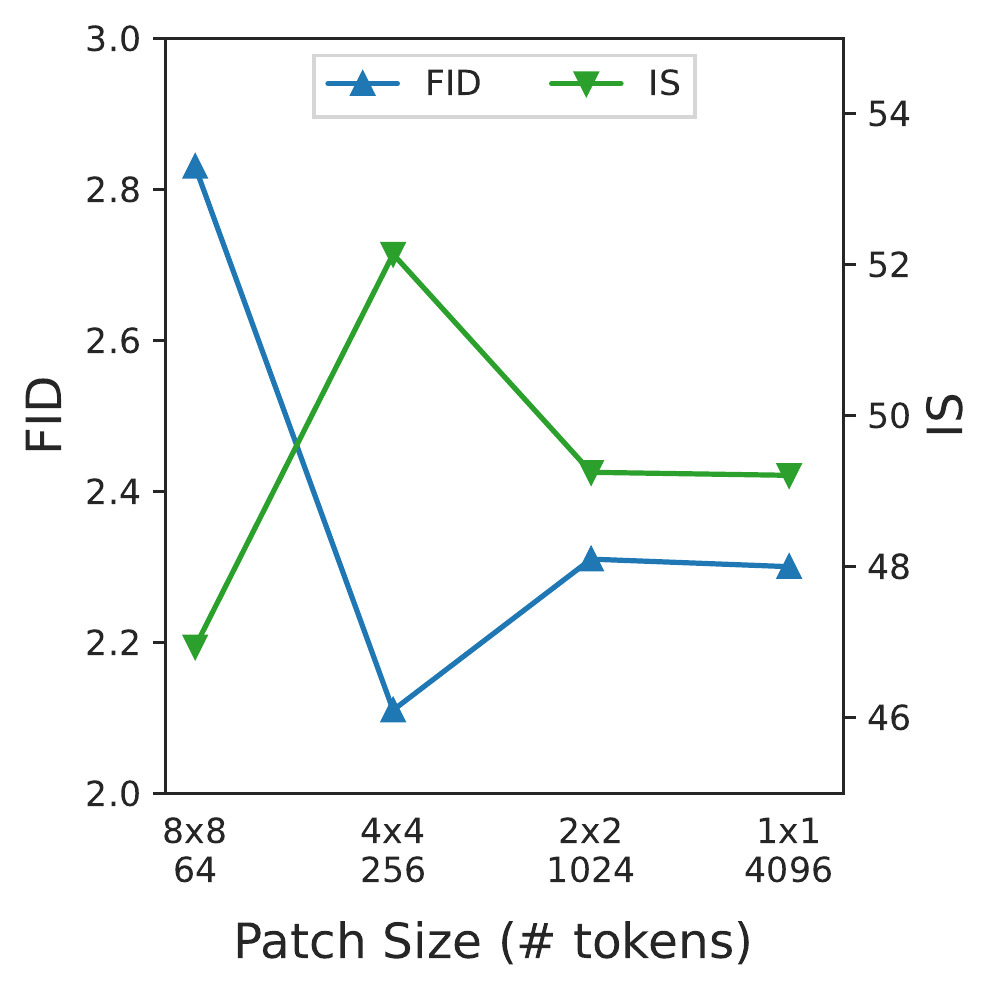}
     \caption{\label{fig:patchsize_ablation}Tokenization}
    \end{subfigure}
    \vspace{-2mm}
    \caption{\textbf{Ablations.} (a) {Effect of the self-conditioning rate for training:} self-conditioning is crucial; a rate of 0 is the special case of no self-conditioning. (b) Effect of the read-write/routing frequency: multiple rounds of read-writes are important to obtain the best result. (c) Effect of tokenization: the model can handle a large number (4096, with $1\by 1$ patches in this case)     of tokens on the inferface.
    }
\label{ablations}
    \vspace{-3mm}
\end{figure*}

\textbf{Image Generation.}
Table~\ref{tab:image_sota} compares our architectures against existing state-of-the-art pixel-space diffusion models on ImageNet.
Despite being fully attention-based and single-scale, our model attains superior generation quality (in both FID and IS) compared to existing models that rely on specialized convolutional architectures, cascaded generation, and/or class-guidance. Both the parameter count and FLOPs are significantly reduced in our model compared to baselines, which is useful for training performant models at higher resolutions without relying on cascades (see samples in Appendix Fig.~\ref{fig:in256_app_1}~\&~\ref{fig:in256_app_2}). For large images ($512$ and $1024$), we report performance of \modelname trained with input scaling \cite{chen2023noise}. We find that 256 latents are sufficient for strong performance even for $1024\by 1024$ images, which produce $16384$ tokens; this is $2\by$ more efficient than the $256\by 256$ ADM UNet, despite operating at $4\by$ higher resolution.

Despite the lack of inductive bias, the model also works well with small datasets such as CIFAR-10. Compared to state-of-the-art FID of 1.79  EDM~\cite{karras2022elucidating}, we obtain 1.81 FID without using their improved sampling procedure. We used a model with 31M params (2x smaller) and trained in 3 hours (10x less) using comparable compute.

\begin{table}[t]
\vspace{-3mm}
\caption{Video Prediction on Kinetics. $\dagger$: reconstruction guidance. 
1: \cite{clark2019adversarial}, 
2: \cite{walker2021predicting},
3: \cite{luc2020transformation},
4: \cite{nash2022transframer},
5: \cite{ho-videodiffusion}.
}
\label{tab:video_sota}
\vspace{-2mm}
\begin{center}
\begin{small}
\begin{tabular}{lcccc}
\toprule
Method & FVD & IS & GFLOPs & Param (M)\\
\midrule
DVD-GAN-FP$^1$ & 69.1 & -- & -- & --\\
Video VQ-VAE$^2$ & 64.3 & -- & -- & -- \\
TrIVD-GAN-FP$^3$ & 25.7 & 12.54 & -- & -- \\
Transframer$^4$ & 25.4 & -- & -- & --\\ 
Video Diffusion$^5\dagger$  & 16.6 & 15.64 &  4136  & 1100\\
\midrule
RIN -- 400 steps  & 11.5 & 17.7 &  386  & 411 \\
RIN -- 1000 steps  & \textbf{10.8} & \textbf{17.7} &  386  & 411 \\
\bottomrule
\end{tabular}
\end{small}
\end{center}
\vspace{-5mm}
\end{table}

\textbf{Video Generation.}
Table~\ref{tab:video_sota} compares our model to existing methods on the Kinetics-600 Video Prediction benchmark.  We follow common practice and use $5$ conditioning frames.  Despite the architecture's simplicity,  \modelname attain superior quality and are more efficient (up to 10$\times$ per step),  without using guidance. Beyond using 3D patches instead of 2D patches, the architecture is identical to that used in $256\by 256$ image generation; while the number of tokens is $2048$, the model can attain strong performance with $256$ latents. The model is especially suitable for video given the intense temporal redundancy, and learns to copy information and dedicate computation to regions of change, as discussed in Section~\ref{sec:visualizing}. Samples can be found in Appendix Fig.~\ref{fig:kinetics_app_1}.

\subsection{Ablations}
For efficiency, we ablate using smaller architectures (latent dimension of 768 instead of 1024) on the ImageNet $64\by 64$ and $128\by128$ tasks with higher learning rate ($2\by10^{-3}$) and fewer updates ($150$k and $220$k, respectively). While this performs worse than our best models, it is sufficient for demonstrating the effect of different design choices.

\textbf{Latent Self-conditioning.}
We study the effect of the rate of self-conditioning at training time. A rate of 0 denotes the special case where no self-conditioning is used (for training nor inference), while a rate $> 0$ e.g. 0.9 means that self-conditioning is used for 90\% of each batch of training tasks (and always used at inference).
As demonstrated in Figure~\ref{fig:sc_rate}, there is a clear correlation between self-conditioning rate and sample quality (i.e., FID/IS), validating the importance using latent self-conditioning to provide context for enhanced routing.
We use $0.9$ for the best results reported.

\textbf{Stacking Blocks.} An important design choice in our architecture is the stacking of blocks to enhance routing. 
For a fair comparison, we analyze the effect of model size on generation quality for a variety of read-write frequencies (Fig.~\ref{fig:tradeoff_read_write}) obtained by stacking blocks with varying $K$ processing layers per block. 
Note that a \textit{single} read-write operation \textit{without latent self-conditioning} is similar to architectures such as PerceiverIO~\cite{jaegle2021perceiverio}.
With a single read-write, the performance saturates earlier as we increase model size. With more frequent read-writes, the model saturates later and with significantly better sample quality, validating the importance of iterative routing.

\textbf{Tokenization.} Recall that images are split into patches to form tokens on the interface. Fig.~\ref{fig:patchsize_ablation} shows that \modelname can handle a wide range of patch sizes. For instance, it can scale to a large number of tokens (4096, for 1$\times$1). While larger patch sizes force tokens to represent more information (i.e., with $8\by 8$ patches), performance remains reasonable.

\textbf{Effect of Noise Schedule.} We find that the sigmoid schedule with an appropriate temperature is more stable training than the cosine schedule, particularly for larger images. For sampling, the noise schedule has less impact and the default cosine schedule can suffice (see Appendix Figure~\ref{fig:noise_ablation}).

\subsection{Visualizing Adaptive Computation}
\label{sec:visualizing}
To better understand the network's emergent adaptive computation, we analyze how information is routed by visualizing the attention distribution of read operations. For image generation, this reveals which parts of the image are most attended to for latent computation. 
Figure~\ref{fig:attn_vis} shows the progression of samples across the reverse process and the read attention (averaged over latents) through the blocks of the corresponding forward pass. 
As the generation progresses, 
the read attention distribution becomes more sparse and favour regions of high information. Since the read attention loads information into the latents for high capacity computation, this suggests that the model learns to dynamically allocate computation on information as needed. More examples for ImageNet can be found in Appendix Fig.~\ref{fig:in256_attn_1}. Appendix Fig.~\ref{fig:kin_attn_1} further shows similar phenomena in the video prediction setting, with the added effect of reading favouring information that cannot merely be copied from conditioning frames, such as object motion and panning.
 \vspace{-1mm}
\section{Related Work}
\vspace{-1mm}
\paragraph{Neural architectures.}
\fullname bear resemblance to architectures that leverage auxiliary memory to decouple computation from the input structure such as Memory Networks~\cite{weston2014memory, sukhbaatar2015end}, Neural Turing Machines~\cite{graves2014neural}, StackRNN~\cite{joulin2015inferring}, Set Transformer~\cite{lee2019set}, Memory Transformers~\cite{burtsev2020memory}, Slot Attention~\cite{locatello2020object}, BigBird~\cite{zaheer2020big}, and Workspace models~\cite{goyal2021coordination}. While latents in our work are similar to auxiliary memory in prior work, we allocate the bulk of computation to latents and iteratively write back updates to the interface, rather than treating them simply as auxiliary memory.
\fullname are perhaps most similar to Set Transformers~\cite{lee2018} and Perceivers ~\cite{jaegle2021perceiver,jaegle2021perceiverio}, which also leverage a set of latents for input-agnostic computation. Unlike these approaches, \modelname alternate computation between the interface and latents, 
which is important for processing of information at both local and global levels without resorting to prohibitively many latents. Moreover, in contrast to existing architectures, latent self-conditioning allows \modelname to leverage recurrence; this allows for propagation of routing context along very deep computation graphs to amortize the cost of iterative routing, which is crucial for achieving strong performance.

Other approaches for adaptive computation have mainly explored models with dynamic depth with recurrent networks~\cite{graves2016adaptive, figurnov2017spatially} or sparse computation~\cite{ yin2021adavit}, facing the challenges non-differentiability and dynamic or masked computation graphs. \modelname are able to allocate compute non-uniformly despite having fixed computation graphs and being differentiable. \modelname are closely related to recurrent models with input attention such as \cite{gregor2015draw}, but scale better by leveraging piecewise optimization enabled by diffusion models.

\vspace{-2mm}
\paragraph{Diffusion Models.} Common diffusion models for images and videos can be roughly divided into pixel diffusion models~\cite{sohl2015deep,ho2020denoising,song2020denoising,dhariwal2021diffusion,ho2021cascaded, karras2022elucidating} and latent diffusion models~\cite{rombach2022high}. In this work we focus on pixel diffusion models due to their relative simplicity. It is known to be challenging to train pixel diffusion models for high resolution images on ImageNet without guidance~\cite{dhariwal2021diffusion,ho2021classifierfree} or cascades~\cite{ho2021cascaded}. We show how improved architectures can allow for scaling pixel-level diffusion models to such large inputs without guidance and cascades, and we expect some insights to transfer to latent diffusion models~\cite{rombach2022high}.

The U-Net~\cite{ronneberger2015u,ho2020denoising} is the predominant architecture for image and video diffusion models~\cite{dhariwal2021diffusion,ho2021cascaded,ho-videodiffusion}. While recent work~\cite{luhman2022improving} has explored pixel-level diffusion with Transformers, they have not been shown to attain strong performance or scale to large inputs. Concurrent work~\cite{peebles2022dit} has shown Transformers may be more tenable when combined with latent diffusion i.e. by downsampling inputs with large-scale pretrained VAEs, but reliance on uniform computation limits gracefully scaling to larger data. Our model suggests a path forward for simple performant and scalable iterative generation of images and video, comparing favourably to U-Nets in sample quality and efficiency, while based on domain-agnostic operations such as attention and fully-connected MLPs, and therefore more universal.

Self-conditioning for diffusion models was originally proposed in~\cite{chen2022analog}. It bears similarity to step-unrolled autoencoders~\cite{savinov2021step}
and has been adopted in several existing work~\cite{strudel2022self,dieleman2022continuous,chen2022generalist}.  While these works condition on predictions of data, latent self-conditioning conditions a neural network on its own hidden activations, akin to recurrent neural network at inference while training without backpropagation through time.
 \section{Conclusion}

\fullname are 
neural networks that explicitly partition hidden units into interface and latent tokens. The interface links the input space to the core computation units operating on the latents, decoupling computation from data layout and allowing adaptive allocation of capacity to different parts of the input.
We show how the challenge of building latents can be amortized in recurrent computation settings -- where the effective network is deep and persistent context can be leveraged -- while still allowing for efficient training. While \modelname are domain-agnostic, we found them to be performant and efficient for image and video generation tasks. As we look towards building more powerful generative models, we hope \modelname can serve as a simple and unified architecture that scales to high-dimensional data across a range of modalities.
To further improve \modelname, we hope to better understand and enhance the effect of latent self-conditioning. Moreover, we hope to combine the advantages of \modelname with orthogonal techniques, such as guidance and latent diffusion. 
\section*{Acknowledgements}
We thank Geoffrey Hinton, Thomas Kipf, Sara Sabour, and Ilija Radosavovic for helpful discussions, and Ben Poole for thoughtful feedback on our draft. We also thank Lala Li, Saurabh Saxena, Ruixiang Zhang for their contributions to the Pix2Seq ~\cite{chen2021pix2seq,chen2022unified} codebase used in this project.

\FloatBarrier
{\small
\bibliography{content/ref}

\begin{thebibliography}{62}
\providecommand{\natexlab}[1]{#1}
\providecommand{\url}[1]{\texttt{#1}}
\expandafter\ifx\csname urlstyle\endcsname\relax
  \providecommand{\doi}[1]{doi: #1}\else
  \providecommand{\doi}{doi: \begingroup \urlstyle{rm}\Url}\fi

\bibitem[Abadi et~al.(2016)Abadi, Barham, Chen, Chen, Davis, Dean, Devin,
  Ghemawat, Irving, Isard, et~al.]{abadi2016tensorflow}
Abadi, M., Barham, P., Chen, J., Chen, Z., Davis, A., Dean, J., Devin, M.,
  Ghemawat, S., Irving, G., Isard, M., et~al.
\newblock $\{$TensorFlow$\}$: a system for $\{$Large-Scale$\}$ machine
  learning.
\newblock In \emph{12th USENIX symposium on operating systems design and
  implementation (OSDI 16)}, pp.\  265--283, 2016.

\bibitem[Ba et~al.(2016)Ba, Kiros, and Hinton]{ba2016layer}
Ba, J.~L., Kiros, J.~R., and Hinton, G.~E.
\newblock Layer normalization.
\newblock \emph{arXiv preprint arXiv:1607.06450}, 2016.

\bibitem[Burtsev et~al.(2020)Burtsev, Kuratov, Peganov, and
  Sapunov]{burtsev2020memory}
Burtsev, M.~S., Kuratov, Y., Peganov, A., and Sapunov, G.~V.
\newblock Memory transformer.
\newblock \emph{arXiv preprint arXiv:2006.11527}, 2020.

\bibitem[Carreira et~al.(2018)Carreira, Noland, Banki-Horvath, Hillier, and
  Zisserman]{carreira2018short}
Carreira, J., Noland, E., Banki-Horvath, A., Hillier, C., and Zisserman, A.
\newblock A short note about kinetics-600.
\newblock \emph{arXiv preprint arXiv:1808.01340}, 2018.

\bibitem[Chen(2023)]{chen2023noise}
Chen, T.
\newblock On the importance of noise schedules for diffusion models.
\newblock \emph{arXiv preprint arXiv:2301.10972}, 2023.

\bibitem[Chen et~al.(2021)Chen, Saxena, Li, Fleet, and Hinton]{chen2021pix2seq}
Chen, T., Saxena, S., Li, L., Fleet, D.~J., and Hinton, G.
\newblock Pix2seq: A language modeling framework for object detection.
\newblock \emph{arXiv preprint arXiv:2109.10852}, 2021.

\bibitem[Chen et~al.(2022{\natexlab{a}})Chen, Li, Saxena, Hinton, and
  Fleet]{chen2022generalist}
Chen, T., Li, L., Saxena, S., Hinton, G., and Fleet, D.~J.
\newblock A generalist framework for panoptic segmentation of images and
  videos.
\newblock \emph{arXiv preprint arXiv:2210.06366}, 2022{\natexlab{a}}.

\bibitem[Chen et~al.(2022{\natexlab{b}})Chen, Saxena, Li, Lin, Fleet, and
  Hinton]{chen2022unified}
Chen, T., Saxena, S., Li, L., Lin, T.-Y., Fleet, D.~J., and Hinton, G.
\newblock A unified sequence interface for vision tasks.
\newblock \emph{arXiv preprint arXiv:2206.07669}, 2022{\natexlab{b}}.

\bibitem[Chen et~al.(2022{\natexlab{c}})Chen, Zhang, and
  Hinton]{chen2022analog}
Chen, T., Zhang, R., and Hinton, G.
\newblock Analog bits: Generating discrete data using diffusion models with
  self-conditioning.
\newblock \emph{arXiv preprint arXiv:2208.04202}, 2022{\natexlab{c}}.

\bibitem[Clark et~al.(2019)Clark, Donahue, and Simonyan]{clark2019adversarial}
Clark, A., Donahue, J., and Simonyan, K.
\newblock Adversarial video generation on complex datasets.
\newblock \emph{arXiv preprint arXiv:1907.06571}, 2019.

\bibitem[Dai et~al.(2019)Dai, Yang, Yang, Carbonell, Le, and
  Salakhutdinov]{dai2019transformer}
Dai, Z., Yang, Z., Yang, Y., Carbonell, J., Le, Q.~V., and Salakhutdinov, R.
\newblock Transformer-xl: Attentive language models beyond a fixed-length
  context.
\newblock \emph{arXiv preprint arXiv:1901.02860}, 2019.

\bibitem[Dhariwal \& Nichol(2022)Dhariwal and Nichol]{dhariwal2021diffusion}
Dhariwal, P. and Nichol, A.
\newblock Diffusion models beat {GAN}s on image synthesis.
\newblock In \emph{{NeurIPS}}, 2022.

\bibitem[Dieleman et~al.(2022)Dieleman, Sartran, Roshannai, Savinov, Ganin,
  Richemond, Doucet, Strudel, Dyer, Durkan, et~al.]{dieleman2022continuous}
Dieleman, S., Sartran, L., Roshannai, A., Savinov, N., Ganin, Y., Richemond,
  P.~H., Doucet, A., Strudel, R., Dyer, C., Durkan, C., et~al.
\newblock Continuous diffusion for categorical data.
\newblock \emph{arXiv preprint arXiv:2211.15089}, 2022.

\bibitem[Dosovitskiy et~al.(2020)Dosovitskiy, Beyer, Kolesnikov, Weissenborn,
  Zhai, Unterthiner, Dehghani, Minderer, Heigold, Gelly,
  et~al.]{dosovitskiy2020image}
Dosovitskiy, A., Beyer, L., Kolesnikov, A., Weissenborn, D., Zhai, X.,
  Unterthiner, T., Dehghani, M., Minderer, M., Heigold, G., Gelly, S., et~al.
\newblock An image is worth 16x16 words: Transformers for image recognition at
  scale.
\newblock \emph{arXiv preprint arXiv:2010.11929}, 2020.

\bibitem[Figurnov et~al.(2017)Figurnov, Collins, Zhu, Zhang, Huang, Vetrov, and
  Salakhutdinov]{figurnov2017spatially}
Figurnov, M., Collins, M.~D., Zhu, Y., Zhang, L., Huang, J., Vetrov, D., and
  Salakhutdinov, R.
\newblock Spatially adaptive computation time for residual networks.
\newblock In \emph{Proceedings of the IEEE conference on computer vision and
  pattern recognition}, pp.\  1039--1048, 2017.

\bibitem[Fukushima(1988)]{fukushima1988neocognitron}
Fukushima, K.
\newblock Neocognitron: A hierarchical neural network capable of visual pattern
  recognition.
\newblock \emph{Neural networks}, 1\penalty0 (2):\penalty0 119--130, 1988.

\bibitem[Goyal et~al.(2021)Goyal, Didolkar, Lamb, Badola, Ke, Rahaman, Binas,
  Blundell, Mozer, and Bengio]{goyal2021coordination}
Goyal, A., Didolkar, A., Lamb, A., Badola, K., Ke, N.~R., Rahaman, N., Binas,
  J., Blundell, C., Mozer, M., and Bengio, Y.
\newblock Coordination among neural modules through a shared global workspace.
\newblock \emph{arXiv preprint arXiv:2103.01197}, 2021.

\bibitem[Graves(2016)]{graves2016adaptive}
Graves, A.
\newblock Adaptive computation time for recurrent neural networks.
\newblock \emph{arXiv preprint arXiv:1603.08983}, 2016.

\bibitem[Graves et~al.(2014)Graves, Wayne, and Danihelka]{graves2014neural}
Graves, A., Wayne, G., and Danihelka, I.
\newblock Neural turing machines.
\newblock \emph{arXiv preprint arXiv:1410.5401}, 2014.

\bibitem[Gregor et~al.(2015)Gregor, Danihelka, Graves, Rezende, and
  Wierstra]{gregor2015draw}
Gregor, K., Danihelka, I., Graves, A., Rezende, D., and Wierstra, D.
\newblock Draw: A recurrent neural network for image generation.
\newblock In \emph{International conference on machine learning}, pp.\
  1462--1471. PMLR, 2015.

\bibitem[He et~al.(2016)He, Zhang, Ren, and Sun]{he2016deep}
He, K., Zhang, X., Ren, S., and Sun, J.
\newblock Deep residual learning for image recognition.
\newblock In \emph{Proceedings of the IEEE conference on computer vision and
  pattern recognition}, pp.\  770--778, 2016.

\bibitem[Hendrycks \& Gimpel(2016)Hendrycks and Gimpel]{hendrycks2016gaussian}
Hendrycks, D. and Gimpel, K.
\newblock Gaussian error linear units (gelus).
\newblock \emph{arXiv preprint arXiv:1606.08415}, 2016.

\bibitem[Heusel et~al.(2017)Heusel, Ramsauer, Unterthiner, Nessler, and
  Hochreiter]{heusel2017gans}
Heusel, M., Ramsauer, H., Unterthiner, T., Nessler, B., and Hochreiter, S.
\newblock Gans trained by a two time-scale update rule converge to a local nash
  equilibrium.
\newblock \emph{Advances in neural information processing systems}, 30, 2017.

\bibitem[Ho \& Salimans(2021)Ho and Salimans]{ho2021classifierfree}
Ho, J. and Salimans, T.
\newblock Classifier-free diffusion guidance.
\newblock In \emph{NeurIPS 2021 Workshop on Deep Generative Models and
  Downstream Applications}, 2021.

\bibitem[Ho et~al.(2020)Ho, Jain, and Abbeel]{ho2020denoising}
Ho, J., Jain, A., and Abbeel, P.
\newblock {Denoising Diffusion Probabilistic Models}.
\newblock \emph{{NeurIPS}}, 2020.

\bibitem[Ho et~al.(2022{\natexlab{a}})Ho, Saharia, Chan, Fleet, Norouzi, and
  Salimans]{ho2021cascaded}
Ho, J., Saharia, C., Chan, W., Fleet, D.~J., Norouzi, M., and Salimans, T.
\newblock Cascaded diffusion models for high fidelity image generation.
\newblock \emph{{JMLR}}, 2022{\natexlab{a}}.

\bibitem[Ho et~al.(2022{\natexlab{b}})Ho, Salimans, Gritsenko, Chan, Norouzi,
  and Fleet]{ho-videodiffusion}
Ho, J., Salimans, T., Gritsenko, A., Chan, W., Norouzi, M., and Fleet, D.~J.
\newblock {Video Diffusion Models}.
\newblock In \emph{NeurIPS}, 2022{\natexlab{b}}.

\bibitem[Jaegle et~al.(2021{\natexlab{a}})Jaegle, Borgeaud, Alayrac, Doersch,
  Ionescu, Ding, Koppula, Zoran, Brock, Shelhamer,
  et~al.]{jaegle2021perceiverio}
Jaegle, A., Borgeaud, S., Alayrac, J.-B., Doersch, C., Ionescu, C., Ding, D.,
  Koppula, S., Zoran, D., Brock, A., Shelhamer, E., et~al.
\newblock Perceiver io: A general architecture for structured inputs \&
  outputs.
\newblock \emph{arXiv preprint arXiv:2107.14795}, 2021{\natexlab{a}}.

\bibitem[Jaegle et~al.(2021{\natexlab{b}})Jaegle, Gimeno, Brock, Vinyals,
  Zisserman, and Carreira]{jaegle2021perceiver}
Jaegle, A., Gimeno, F., Brock, A., Vinyals, O., Zisserman, A., and Carreira, J.
\newblock Perceiver: General perception with iterative attention.
\newblock In \emph{International conference on machine learning}, pp.\
  4651--4664. PMLR, 2021{\natexlab{b}}.

\bibitem[Joulin \& Mikolov(2015)Joulin and Mikolov]{joulin2015inferring}
Joulin, A. and Mikolov, T.
\newblock Inferring algorithmic patterns with stack-augmented recurrent nets.
\newblock \emph{Advances in neural information processing systems}, 28, 2015.

\bibitem[Karras et~al.(2022)Karras, Aittala, Aila, and
  Laine]{karras2022elucidating}
Karras, T., Aittala, M., Aila, T., and Laine, S.
\newblock Elucidating the design space of diffusion-based generative models.
\newblock \emph{arXiv preprint arXiv:2206.00364}, 2022.

\bibitem[Kingma et~al.(2021)Kingma, Salimans, Poole, and
  Ho]{kingma2021variational}
Kingma, D., Salimans, T., Poole, B., and Ho, J.
\newblock Variational diffusion models.
\newblock \emph{Advances in neural information processing systems},
  34:\penalty0 21696--21707, 2021.

\bibitem[Krizhevsky et~al.()Krizhevsky, Nair, and Hinton]{cifar10}
Krizhevsky, A., Nair, V., and Hinton, G.
\newblock Cifar-10 (canadian institute for advanced research).
\newblock URL \url{http://www.cs.toronto.edu/~kriz/cifar.html}.

\bibitem[Krizhevsky et~al.(2012)Krizhevsky, Sutskever, and
  Hinton]{krizhevsky2012imagenet}
Krizhevsky, A., Sutskever, I., and Hinton, G.~E.
\newblock Imagenet classification with deep convolutional neural networks.
\newblock In \emph{Advances in neural information processing systems}, 2012.

\bibitem[LeCun et~al.(1989)LeCun, Boser, Denker, Henderson, Howard, Hubbard,
  and Jackel]{lecun1989backpropagation}
LeCun, Y., Boser, B., Denker, J.~S., Henderson, D., Howard, R.~E., Hubbard, W.,
  and Jackel, L.~D.
\newblock Backpropagation applied to handwritten zip code recognition.
\newblock \emph{Neural computation}, 1\penalty0 (4):\penalty0 541--551, 1989.

\bibitem[Lee et~al.(2018)Lee, Lee, Kim, Kosiorek, Choi, and Teh]{lee2018}
Lee, J., Lee, Y., Kim, J., Kosiorek, A.~R., Choi, S., and Teh, Y.~W.
\newblock Set transformer.
\newblock \emph{CoRR}, abs/1810.00825, 2018.
\newblock URL \url{http://arxiv.org/abs/1810.00825}.

\bibitem[Lee et~al.(2019)Lee, Lee, Kim, Kosiorek, Choi, and Teh]{lee2019set}
Lee, J., Lee, Y., Kim, J., Kosiorek, A., Choi, S., and Teh, Y.~W.
\newblock Set transformer: A framework for attention-based
  permutation-invariant neural networks.
\newblock In \emph{International conference on machine learning}, pp.\
  3744--3753. PMLR, 2019.

\bibitem[Locatello et~al.(2020)Locatello, Weissenborn, Unterthiner, Mahendran,
  Heigold, Uszkoreit, Dosovitskiy, and Kipf]{locatello2020object}
Locatello, F., Weissenborn, D., Unterthiner, T., Mahendran, A., Heigold, G.,
  Uszkoreit, J., Dosovitskiy, A., and Kipf, T.
\newblock Object-centric learning with slot attention.
\newblock \emph{Advances in Neural Information Processing Systems},
  33:\penalty0 11525--11538, 2020.

\bibitem[Luc et~al.(2020)Luc, Clark, Dieleman, Casas, Doron, Cassirer, and
  Simonyan]{luc2020transformation}
Luc, P., Clark, A., Dieleman, S., Casas, D. d.~L., Doron, Y., Cassirer, A., and
  Simonyan, K.
\newblock Transformation-based adversarial video prediction on large-scale
  data.
\newblock \emph{arXiv preprint arXiv:2003.04035}, 2020.

\bibitem[Luhman \& Luhman(2022)Luhman and Luhman]{luhman2022improving}
Luhman, T. and Luhman, E.
\newblock Improving diffusion model efficiency through patching.
\newblock \emph{arXiv preprint arXiv:2207.04316}, 2022.

\bibitem[Nash et~al.(2022)Nash, Carreira, Walker, Barr, Jaegle, Malinowski, and
  Battaglia]{nash2022transframer}
Nash, C., Carreira, J., Walker, J., Barr, I., Jaegle, A., Malinowski, M., and
  Battaglia, P.
\newblock Transframer: Arbitrary frame prediction with generative models.
\newblock \emph{arXiv preprint arXiv:2203.09494}, 2022.

\bibitem[Nichol \& Dhariwal(2021)Nichol and Dhariwal]{nichol2021improved}
Nichol, A. and Dhariwal, P.
\newblock Improved denoising diffusion probabilistic models.
\newblock \emph{arXiv preprint arXiv:2102.09672}, 2021.

\bibitem[Paszke et~al.(2019)Paszke, Gross, Massa, Lerer, Bradbury, Chanan,
  Killeen, Lin, Gimelshein, Antiga, et~al.]{paszke2019pytorch}
Paszke, A., Gross, S., Massa, F., Lerer, A., Bradbury, J., Chanan, G., Killeen,
  T., Lin, Z., Gimelshein, N., Antiga, L., et~al.
\newblock Pytorch: An imperative style, high-performance deep learning library.
\newblock \emph{Advances in neural information processing systems}, 32, 2019.

\bibitem[Peebles \& Xie(2022)Peebles and Xie]{peebles2022dit}
Peebles, W. and Xie, S.
\newblock Scalable diffusion models with transformers.
\newblock \emph{arXiv preprint arXiv:2212.09748}, 2022.

\bibitem[Rae et~al.(2019)Rae, Potapenko, Jayakumar, and
  Lillicrap]{rae2019compressive}
Rae, J.~W., Potapenko, A., Jayakumar, S.~M., and Lillicrap, T.~P.
\newblock Compressive transformers for long-range sequence modelling.
\newblock \emph{arXiv preprint arXiv:1911.05507}, 2019.

\bibitem[Rombach et~al.(2022)Rombach, Blattmann, Lorenz, Esser, and
  Ommer]{rombach2022high}
Rombach, R., Blattmann, A., Lorenz, D., Esser, P., and Ommer, B.
\newblock High-resolution image synthesis with latent diffusion models.
\newblock In \emph{Proceedings of the IEEE/CVF Conference on Computer Vision
  and Pattern Recognition}, pp.\  10684--10695, 2022.

\bibitem[Ronneberger et~al.(2015)Ronneberger, Fischer, and
  Brox]{ronneberger2015u}
Ronneberger, O., Fischer, P., and Brox, T.
\newblock U-net: Convolutional networks for biomedical image segmentation.
\newblock In \emph{International Conference on Medical image computing and
  computer-assisted intervention}, pp.\  234--241. Springer, 2015.

\bibitem[Russakovsky et~al.(2015)Russakovsky, Deng, Su, Krause, Satheesh, Ma,
  Huang, Karpathy, Khosla, Bernstein, et~al.]{russakovsky2015imagenet}
Russakovsky, O., Deng, J., Su, H., Krause, J., Satheesh, S., Ma, S., Huang, Z.,
  Karpathy, A., Khosla, A., Bernstein, M., et~al.
\newblock Imagenet large scale visual recognition challenge.
\newblock \emph{International journal of computer vision}, 115\penalty0
  (3):\penalty0 211--252, 2015.

\bibitem[Salimans et~al.(2016)Salimans, Goodfellow, Zaremba, Cheung, Radford,
  and Chen]{salimans2016improved}
Salimans, T., Goodfellow, I., Zaremba, W., Cheung, V., Radford, A., and Chen,
  X.
\newblock Improved techniques for training gans.
\newblock \emph{Advances in neural information processing systems}, 29, 2016.

\bibitem[Savinov et~al.(2021)Savinov, Chung, Binkowski, Elsen, and
  Oord]{savinov2021step}
Savinov, N., Chung, J., Binkowski, M., Elsen, E., and Oord, A. v.~d.
\newblock Step-unrolled denoising autoencoders for text generation.
\newblock \emph{arXiv preprint arXiv:2112.06749}, 2021.

\bibitem[Sohl-Dickstein et~al.(2015)Sohl-Dickstein, Weiss, Maheswaranathan, and
  Ganguli]{sohl2015deep}
Sohl-Dickstein, J., Weiss, E., Maheswaranathan, N., and Ganguli, S.
\newblock Deep unsupervised learning using nonequilibrium thermodynamics.
\newblock In \emph{International Conference on Machine Learning}, pp.\
  2256--2265. PMLR, 2015.

\bibitem[Song et~al.(2020)Song, Meng, and Ermon]{song2020denoising}
Song, J., Meng, C., and Ermon, S.
\newblock Denoising diffusion implicit models.
\newblock \emph{arXiv preprint arXiv:2010.02502}, 2020.

\bibitem[Song et~al.(2021)Song, Sohl-Dickstein, Kingma, Kumar, Ermon, and
  Poole]{song2021scorebased}
Song, Y., Sohl-Dickstein, J., Kingma, D.~P., Kumar, A., Ermon, S., and Poole,
  B.
\newblock Score-based generative modeling through stochastic differential
  equations.
\newblock In \emph{International Conference on Learning Representations}, 2021.

\bibitem[Strudel et~al.(2022)Strudel, Tallec, Altch{\'e}, Du, Ganin, Mensch,
  Grathwohl, Savinov, Dieleman, Sifre, et~al.]{strudel2022self}
Strudel, R., Tallec, C., Altch{\'e}, F., Du, Y., Ganin, Y., Mensch, A.,
  Grathwohl, W., Savinov, N., Dieleman, S., Sifre, L., et~al.
\newblock Self-conditioned embedding diffusion for text generation.
\newblock \emph{arXiv preprint arXiv:2211.04236}, 2022.

\bibitem[Sukhbaatar et~al.(2015)Sukhbaatar, Weston, Fergus,
  et~al.]{sukhbaatar2015end}
Sukhbaatar, S., Weston, J., Fergus, R., et~al.
\newblock End-to-end memory networks.
\newblock \emph{Advances in neural information processing systems}, 28, 2015.

\bibitem[Unterthiner et~al.(2018)Unterthiner, van Steenkiste, Kurach, Marinier,
  Michalski, and Gelly]{unterthiner2018towards}
Unterthiner, T., van Steenkiste, S., Kurach, K., Marinier, R., Michalski, M.,
  and Gelly, S.
\newblock Towards accurate generative models of video: A new metric \&
  challenges.
\newblock \emph{arXiv preprint arXiv:1812.01717}, 2018.

\bibitem[Vaswani et~al.(2017)Vaswani, Shazeer, Parmar, Uszkoreit, Jones, Gomez,
  Kaiser, and Polosukhin]{vaswani2017attention}
Vaswani, A., Shazeer, N., Parmar, N., Uszkoreit, J., Jones, L., Gomez, A.~N.,
  Kaiser, {\L}., and Polosukhin, I.
\newblock Attention is all you need.
\newblock \emph{Advances in neural information processing systems}, 30, 2017.

\bibitem[Walker et~al.(2021)Walker, Razavi, and Oord]{walker2021predicting}
Walker, J., Razavi, A., and Oord, A. v.~d.
\newblock Predicting video with vqvae.
\newblock \emph{arXiv preprint arXiv:2103.01950}, 2021.

\bibitem[Weston et~al.(2014)Weston, Chopra, and Bordes]{weston2014memory}
Weston, J., Chopra, S., and Bordes, A.
\newblock Memory networks.
\newblock \emph{arXiv preprint arXiv:1410.3916}, 2014.

\bibitem[Yin et~al.(2021)Yin, Vahdat, Alvarez, Mallya, Kautz, and
  Molchanov]{yin2021adavit}
Yin, H., Vahdat, A., Alvarez, J., Mallya, A., Kautz, J., and Molchanov, P.
\newblock Adavit: Adaptive tokens for efficient vision transformer.
\newblock \emph{arXiv preprint arXiv:2112.07658}, 2021.

\bibitem[You et~al.(2019)You, Li, Reddi, Hseu, Kumar, Bhojanapalli, Song,
  Demmel, Keutzer, and Hsieh]{you2019large}
You, Y., Li, J., Reddi, S., Hseu, J., Kumar, S., Bhojanapalli, S., Song, X.,
  Demmel, J., Keutzer, K., and Hsieh, C.-J.
\newblock Large batch optimization for deep learning: Training bert in 76
  minutes.
\newblock \emph{arXiv preprint arXiv:1904.00962}, 2019.

\bibitem[Zaheer et~al.(2020)Zaheer, Guruganesh, Dubey, Ainslie, Alberti,
  Ontanon, Pham, Ravula, Wang, Yang, et~al.]{zaheer2020big}
Zaheer, M., Guruganesh, G., Dubey, K.~A., Ainslie, J., Alberti, C., Ontanon,
  S., Pham, P., Ravula, A., Wang, Q., Yang, L., et~al.
\newblock Big bird: Transformers for longer sequences.
\newblock \emph{Advances in Neural Information Processing Systems},
  33:\penalty0 17283--17297, 2020.

\end{thebibliography}
\bibliographystyle{icml2023}}

\cleardoublepage
\appendix
\counterwithin{figure}{section}
\counterwithin{table}{section}
\onecolumn

\appendix
\section{Architecture Implementation Pseudo-code}
\label{sec:net_implementation}

Algorithm~\ref{alg:net_implementation} provides a more detailed implementation of \modelname. Note that for clarity, we only show it for image generation task, but for other tasks or data modalities, we only need to change the interface initialization, i.e. the tokenization of the input. We also omit some functions, such as ``multihead\_attention'' and ``ffn'' (i.e. feed-forward network), which are specified in Transformers~\cite{vaswani2017attention} and available as APIs in major deep learning frameworks, such as Tensorflow~\cite{abadi2016tensorflow} and PyTorch~\cite{paszke2019pytorch}.

\begin{figure}[ht]
  \centering
\begin{minipage}[t]{0.7\textwidth}
\begin{algorithm}[H]
\small
\caption{\modelname Implementation Pseudo-code.}
\label{alg:net_implementation}
\definecolor{codeblue}{rgb}{0.25,0.5,0.5}
\definecolor{codekw}{rgb}{0.85, 0.18, 0.50}
\lstset{
  backgroundcolor=\color{white},
  basicstyle=\fontsize{7.5pt}{7.5pt}\ttfamily\selectfont,
  columns=fullflexible,
  breaklines=true,
  captionpos=b,
  commentstyle=\fontsize{7.5pt}{7.5pt}\color{codeblue},
  keywordstyle=\fontsize{7.5pt}{7.5pt}\color{codekw},
  escapechar={|}, 
}
\begin{lstlisting}[language=python]
def block(z, x, num_layers):
  """Core computation block."""
  z = z + multihead_attention(q=layer_norm(z), kv=x, n_heads=16)
  z = z + ffn(layer_norm(z), expansion=4)
  
  for _ in range(num_layers):
    zn = layer_norm(z)
    z = z + multihead_attention(q=zn, kv=zn, n_heads=16)
    z = z + ffn(layer_norm(z), expansion=4)
  
  x = x + multihead_attention(q=layer_norm(x), kv=z, n_heads=16)
  x = x + ffn(layer_norm(x), expansion=4)
  
  return z, x


def rin(x, patch_size, num_latents, latent_dim, interface_dim, 
                        num_blocks, num_layers_per_block, prev_latents=None):
  """Forward pass of Network."""
  bsz, image_size, _, _ = x.shape
  size = image_size // patch_size

  # Initialize interface (with image tokenization as an example)
  x = conv(x, kernel_size=patch_size, stride=patch_size, padding='SAME')
  pos_emb = truncated_normal((1, size, size, dim), scale=0.02)
  x = layer_norm(x) + pos_emb

  
  # Initialize latents
  z = truncated_normal((num_latents, latent_dim), scale=0.02)
  
  # Latent self-conditioning
  if prev_latents is not None:
    prev_latents = prev_latents + ffn(stop_grad(prev_latents), expansion=4)
    z = z + layer_norm(prev_latents, init_scale=0, init_bias=0)
  
  # Compute
  for _ in range(num_blocks):
    z, x = block(z, x, num_layers_per_block)
  
  # Readout
  x = linear(layer_norm(x), dim=3*patch_size**2)
  x = depth_to_space(reshape(x, [bsz, size, size, -1]), patch_size)
  
  return z, x    
  
\end{lstlisting}
\end{algorithm}
\end{minipage}
\end{figure}

\newpage
\section{More Details of Training / Sampling Algorithms, and Noise schedules}
\label{sec:alg_details}

Algorithm~\ref{alg:gamma} contains different choices of $\gamma(t)$, the continuous time noise schedule function.
\vspace{-4mm}
\begin{figure}[ht]
  \centering
\begin{minipage}[t]{0.8\textwidth}
\begin{algorithm}[H]
\small
\caption{\small Continuous time noise scheduling function.
}
\label{alg:gamma}
\definecolor{codeblue}{rgb}{0.25,0.5,0.5}
\definecolor{codekw}{rgb}{0.85, 0.18, 0.50}
\lstset{
  backgroundcolor=\color{white},
  basicstyle=\fontsize{7.5pt}{7.5pt}\ttfamily\selectfont,
  columns=fullflexible,
  breaklines=true,
  captionpos=b,
  commentstyle=\fontsize{7.5pt}{7.5pt}\color{codeblue},
  keywordstyle=\fontsize{7.5pt}{7.5pt}\color{codekw},
  escapechar={|}, 
}
\begin{lstlisting}[language=python]
def gamma_cosine_schedule(t, ns=0.0002, ds=0.00025):
  # A scheduling function based on cosine function.
  return numpy.cos(((t + ns) / (1 + ds)) * numpy.pi / 2)**2
  
def gamma_sigmoid_schedule(t, start=-3, end=3, tau=1.0, clip_min=1e-9):
  # A scheduling function based on sigmoid function.
  v_start = sigmoid(start / tau)
  v_end = sigmoid(end / tau)
  output = (-sigmoid((t * (end - start) + start) / tau) + v_end) / (v_end - v_start)
  return np.clip(output, clip_min, 1.)
\end{lstlisting}
\end{algorithm}
\end{minipage}
\end{figure}

Algorithm~\ref{alg:ddim_ddpm} contains DDIM~\cite{song2020denoising} and DDPM~\cite{ho2020denoising} updating rules, as specified in~\cite{chen2022analog}.
\vspace{-4mm}
\begin{figure}[ht]
  \centering
\begin{minipage}[t]{0.6\textwidth}
\begin{algorithm}[H]
\small
\caption{\small $x_t$ estimation with DDIM / DDPM updating rules.
}
\label{alg:ddim_ddpm}
\definecolor{codeblue}{rgb}{0.25,0.5,0.5}
\definecolor{codekw}{rgb}{0.85, 0.18, 0.50}
\lstset{
  backgroundcolor=\color{white},
  basicstyle=\fontsize{7.5pt}{7.5pt}\ttfamily\selectfont,
  columns=fullflexible,
  breaklines=true,
  captionpos=b,
  commentstyle=\fontsize{7.5pt}{7.5pt}\color{codeblue},
  keywordstyle=\fontsize{7.5pt}{7.5pt}\color{codekw},
  escapechar={|}, 
}
\begin{lstlisting}[language=python]
def ddim_step(x_t, x_pred, t_now, t_next):
  # Estimate x at t_next with DDIM updating rule.
  |$\gamma_{\text{now}}$| = gamma(t_now)
  |$\gamma_{\text{next}}$| = gamma(t_next)
  x_pred = clip(x_pred, -scale, scale)
  eps =|$\frac{1}{\sqrt{1-\gamma_\text{now}}}$| * (x_t - |$\sqrt{\gamma_\text{now}}$| * x_pred)
  x_next = |$\sqrt{\gamma_\text{next}}$| * x_pred + |$\sqrt{1-\gamma_\text{next}}$| * eps
  return x_next
  
def ddpm_step(x_t, x_pred, t_now, t_next):
  # Estimate x at t_next with DDPM updating rule.
  |$\gamma_{\text{now}}$| = gamma(t_now)
  |$\alpha_{\text{now}}$| = gamma(t_now) / gamma(t_next)
  |$\sigma_{\text{now}}$| = sqrt(1 - |$\alpha_{\text{now}}$|)
  z = normal(mean=0, std=1)
  x_pred = clip(x_pred, -scale, scale)
  eps =|$\frac{1}{\sqrt{1-\gamma_\text{now}}}$| * (x_t - |$\sqrt{\gamma_\text{now}}$| * x_pred)
  x_next = |$\frac{1}{\sqrt{\alpha_{\text{now}}}}$| * (x_t - |$\frac{1 - \alpha_{\text{now}}}{\sqrt{1 - \gamma_{\text{now}}}}$| * eps) + |$\sigma_{\text{now}}$| * z
  return x_next
\end{lstlisting}
\end{algorithm}
\end{minipage}
\vspace{-2mm}
\end{figure}

\begin{figure}[h!]
\begin{center}  
\includegraphics[width=0.6\linewidth]{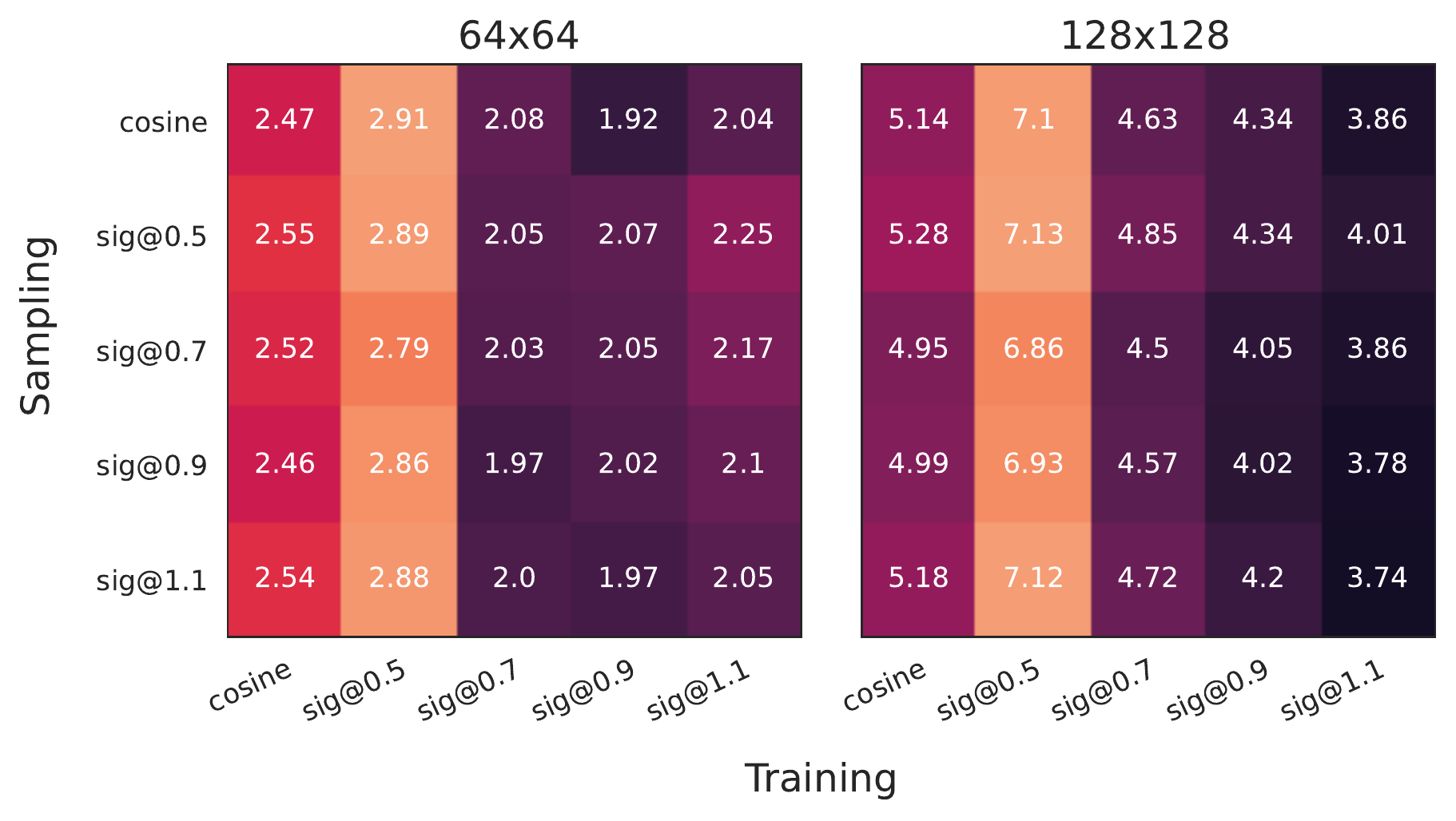}
\end{center}
\vspace{-5mm}
\caption{\label{fig:noise_ablation} \textbf{Effect of noise schedule.} Comparing noise schedules for training and sampling, with corresponding FID score. The sigmoid schedule with an appropriate temperature is more stable during training than the widely used cosine schedule, particularly for larger images. For sampling, the noise schedule has less impact and the default cosine schedule can suffice.
}
\vspace{-3mm}
\end{figure}

\section{Hyper-parameters and Other Training Details}
\label{sec:train_details}
We train most models on 32 TPUv3 chips with a batch size of 1024. Models for $512\by512$ and $1024\by 1024$ are trained on 64 TPUv3 chips and 256 TPUv4 chips, respectively. All models are trained with the LAMB optimizer~\cite{you2019large}.
\vspace{-4mm}
\begin{table}[h]
\caption{Model Hyper-parameters.}
\label{tab:model_hypers}
\begin{center}
\begin{small}
\begin{tabular}{lcccccccccc}
\toprule
Task & Input/Output & Blocks & Depth & Latents & dim(Z) & dim(X) & Tokens (patch size) & Heads & Params & GFLOPs \\
\midrule
IN & $64\by 64 \by 3$  & 4 & 4 & 128 &  1024  & 256 & 256 ($4\by 4$)& 16 & 280M & 106 \\
IN & $128\by 128 \by 3$  & 6 & 4 & 128 &  1024  & 512 & 1024 ($4\by 4$)& 16 & 410M & 194 \\
IN & $256\by 256 \by 3$  & 6 & 4 & 256 &  1024  & 512 & 1024 ($8\by 8$) & 16 & 410M & 334 \\
IN & $512\by 512 \by 3$  & 6 & 6 & 256 &  768  & 512 & 4096 ($8\by 8$) & 16 & 320M & 415 \\
IN & $1024\by 1024 \by 3$  & 6 & 8 & 256 &  768  & 512 & 16384 ($8\by 8$) & 16 & 415M & 1120 \\
K-600 & $16\by 64\by 64 \by 3$  & 6 & 4 & 256 &  1024  & 512 & 2048 ($2\by4\by 4$) & 16 & 411M & 386 \\
\bottomrule
\end{tabular}
\end{small}
\end{center}
\vspace{-4mm}
\end{table}
\begin{table}[h!]
\caption{Training Hyper-parameters.}
\label{tab:train_hypers}
\begin{center}
\begin{small}
\begin{tabular}{lcccccccccc}
\toprule
Task & Input/Output & Updates & Batch Size & LR & LR-decay & Optim $\beta_2$ & Weight Dec. & Self-cond. Rate & EMA $\beta$ \\
\midrule
IN & $64\by 64 \by 3$  & 300K & 1024  & 1e-3 & cosine &  0.999  & 0.01 & 0.9 & 0.9999 \\
IN & $128\by 128 \by 3$  & 600K & 1024  &1e-3 & cosine &  0.999  & 0.001 & 0.9& 0.9999 \\
IN & $256\by 256 \by 3$  & 600K & 1024  &1e-3 & cosine & 0.999  & 0.001 & 0.9& 0.9999 \\
IN & $512\by 512 \by 3$  & 1M & 1024  &1e-3 & cosine & 0.999  & 0.01 & 0.9& 0.9999 \\
IN & $1024\by 1024 \by 3$  & 1M & 512  &1e-3 & None & 0.999  & 0.01 & 0.9& 0.9999 \\
K-600 & $16\by 64\by 64 \by 3$  & 500K & 1024  & 1e-3 & cosine & 0.999 &  0.001  & 0.85 & 0.99 \\
\bottomrule
\end{tabular}
\end{small}
\end{center}
\vskip -0.1in
\end{table}

\newpage
\section{Sample Visualizations}
\label{sec:samples}
\begin{figure}[!htb]
\begin{center}  
\includegraphics[width=0.70\linewidth]{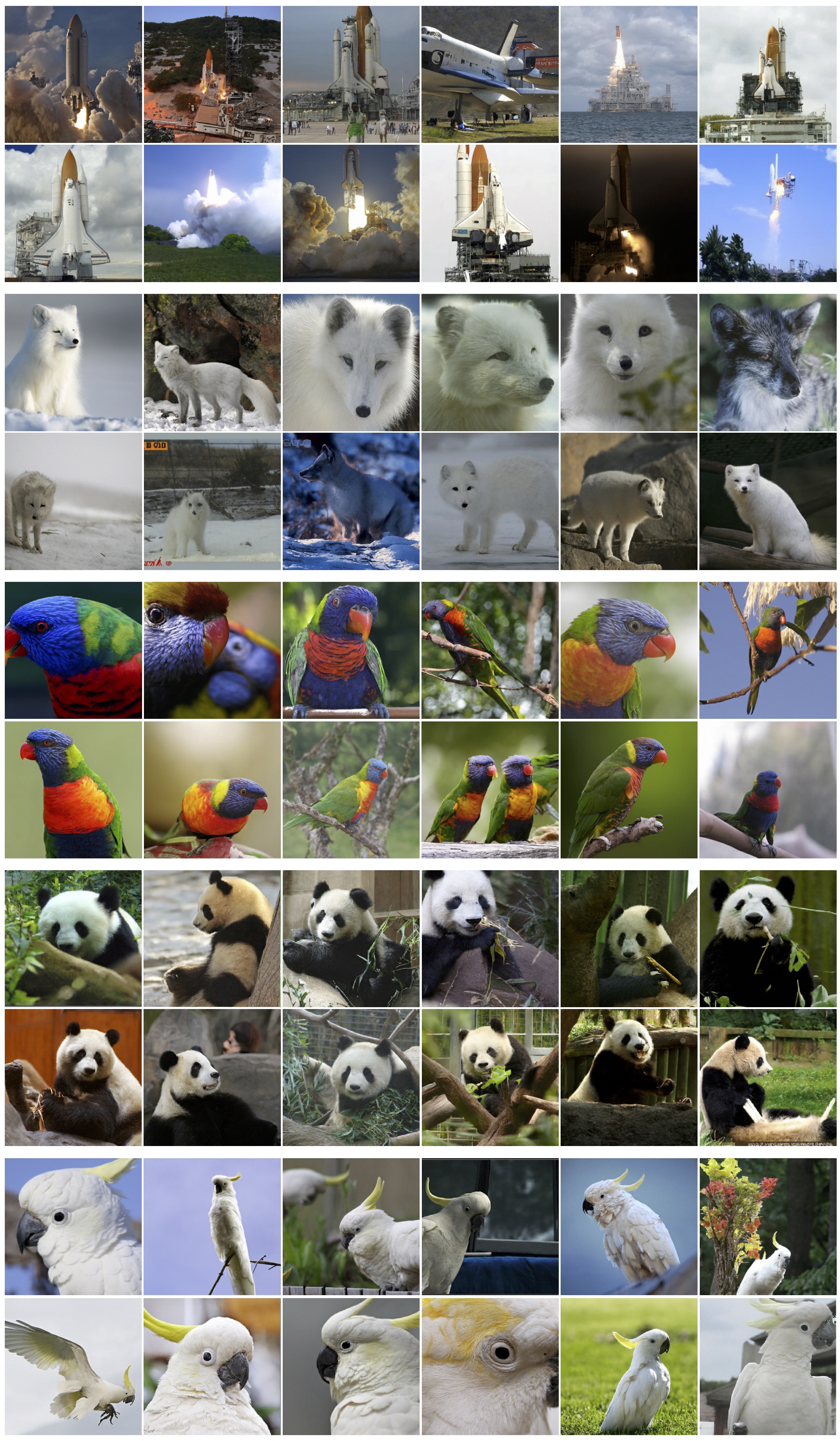}
\end{center}
\caption{\label{fig:in256_app_1} Selected class-conditional samples from a model trained on ImageNet $256\by 256$. Classes from the top: space shuttle (812),
arctic fox (279),
lorikeet (90),
giant panda (388),
cockatoo (89).
}
\end{figure}
\begin{figure}[h!]
\vspace{6mm}
\begin{center}  
\includegraphics[width=0.7\linewidth]{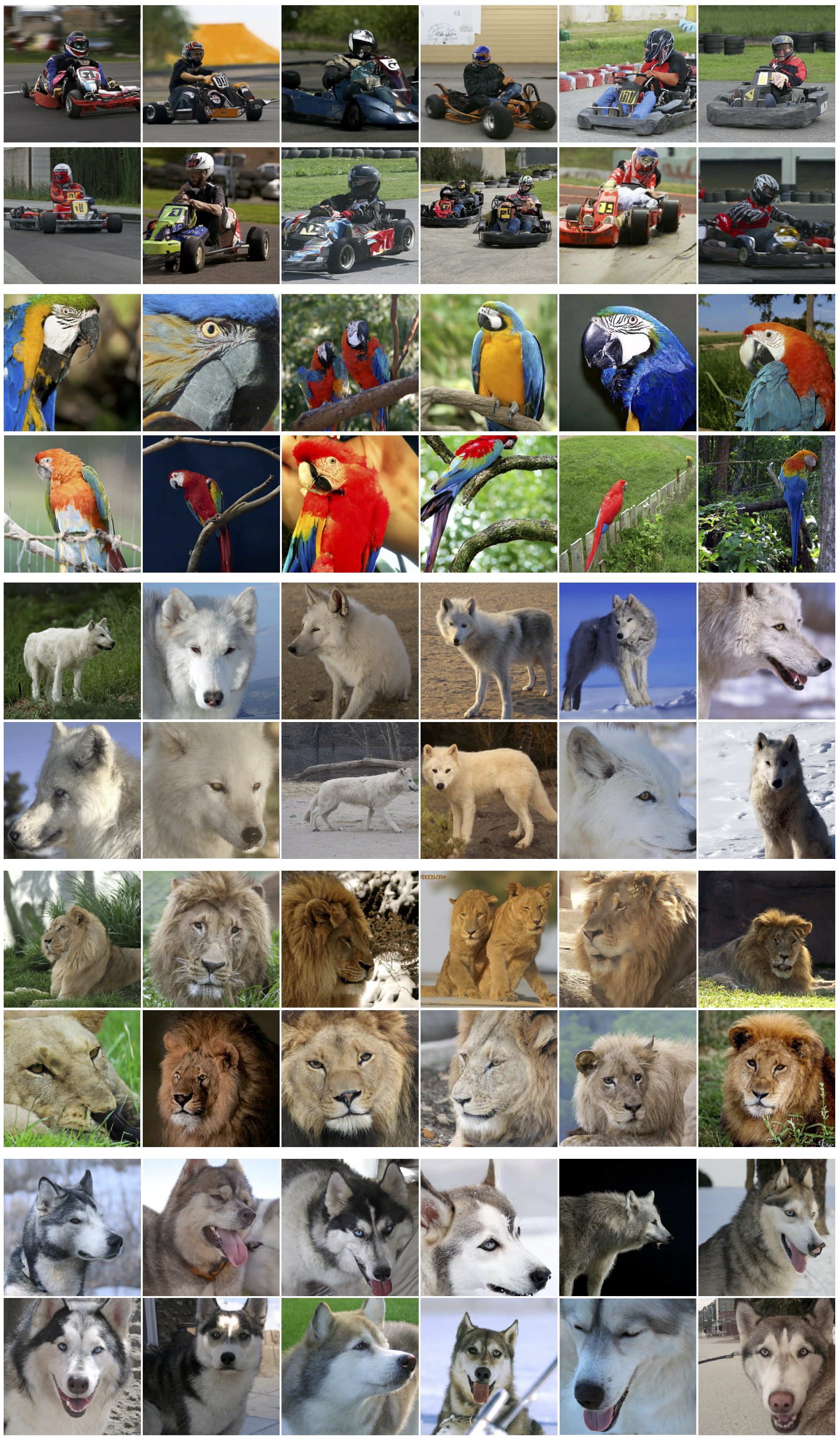}
\end{center}
\caption{\label{fig:in256_app_2} Selected class-conditional samples from a model trained on ImageNet $256\by 256$. Classes from the top: go-kart (573),
macaw (88),
white wolf (270),
lion (291),
siberian husky (250).
}
\end{figure}

\begin{figure}[h!]
\begin{center}  
\includegraphics[width=0.93\linewidth]{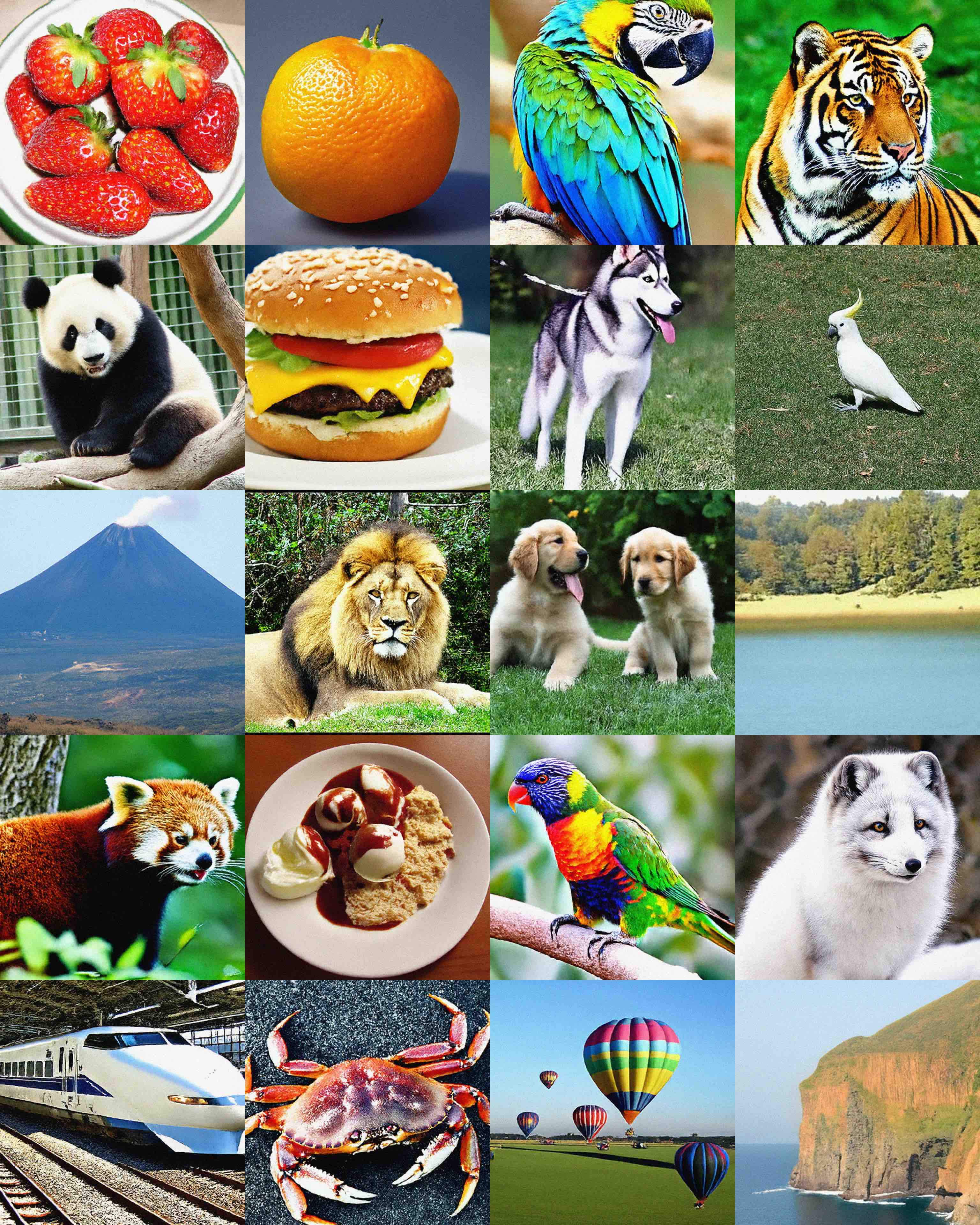}
\end{center}
\caption{\label{fig:in768}Random class-conditional ImageNet $768\by 768$ samples generated by \modelname trained with input scaling \cite{chen2023noise}. Note that these samples are uncurated but generated using classifier-free guidance. These demonstrate the architecture can scale to higher-resolutions despite being single-scale and operating directly on pixels.
}
\end{figure}

\begin{figure}[h!]
\vspace{6mm}
\begin{center}  
\includegraphics[width=0.93\linewidth]{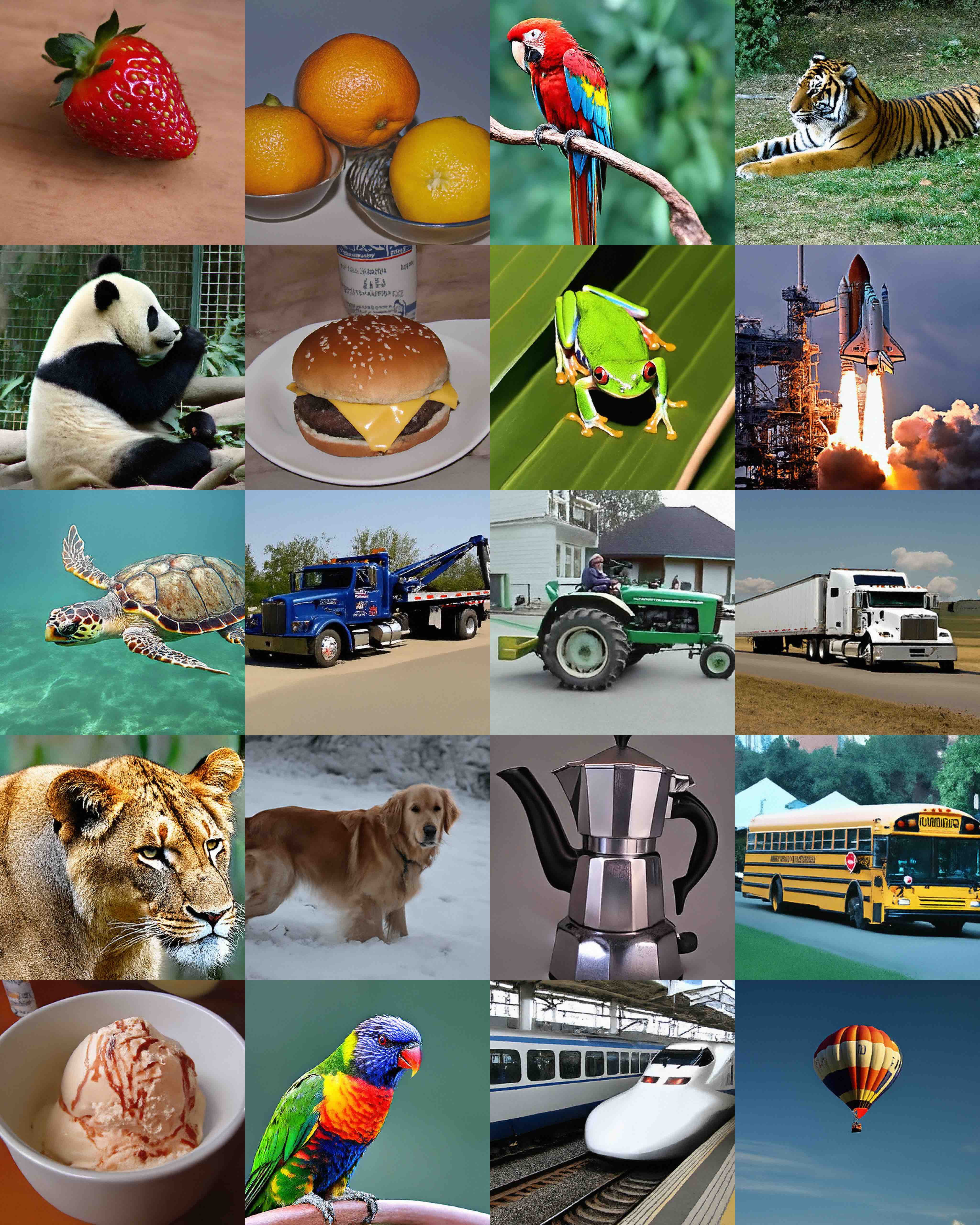}
\end{center}
\caption{\label{fig:in1024}Random class-conditional ImageNet $1024\by 1024$ samples generated by \modelname trained with input scaling \cite{chen2023noise}. Note that these samples are uncurated but generated using classifier-free guidance.
}
\end{figure}

\begin{figure}[h!]
\begin{center}  
\includegraphics[width=0.88\linewidth]{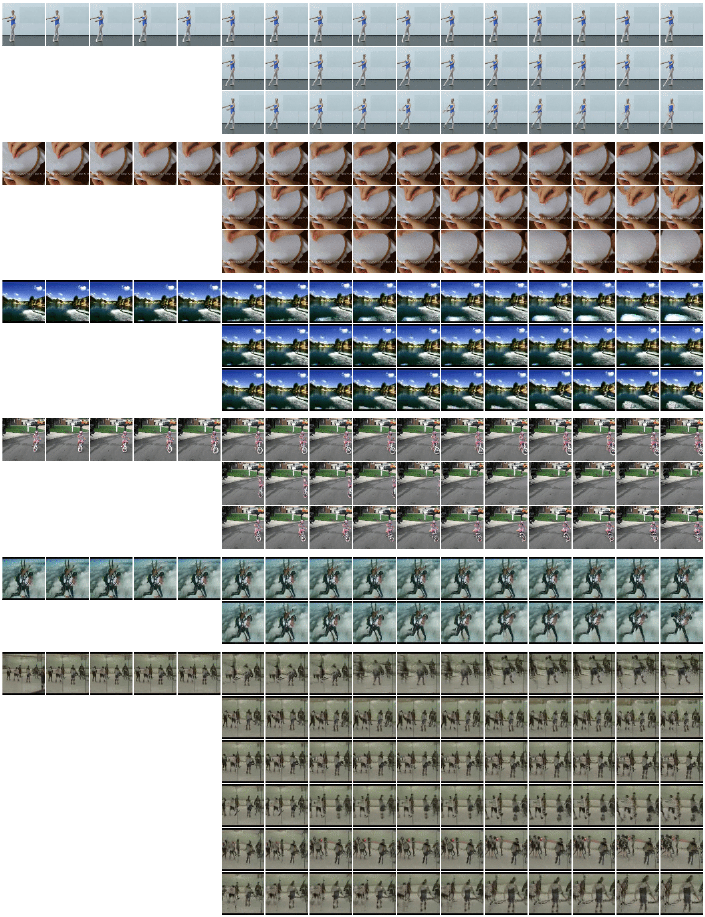}
\end{center}
\caption{\label{fig:kinetics_app_1}Selected samples of video prediction on Kinetics-600 at $16\by 64 \by 64$ showing examples of multi-modality across different future predictions, with conditioning frames from the test set. For example, the ballerina's arm and leg movements vary (first); the hand moves in different ways while sewing (second); the wakeboarder faces different waves (third); the bicyclist takes different turns; the sky-divers face different fates; the hockey scene (last) is zoomed and panned in different ways.
}
\end{figure}

\begin{figure}[h!]
\begin{center}  
\includegraphics[width=0.85\linewidth]{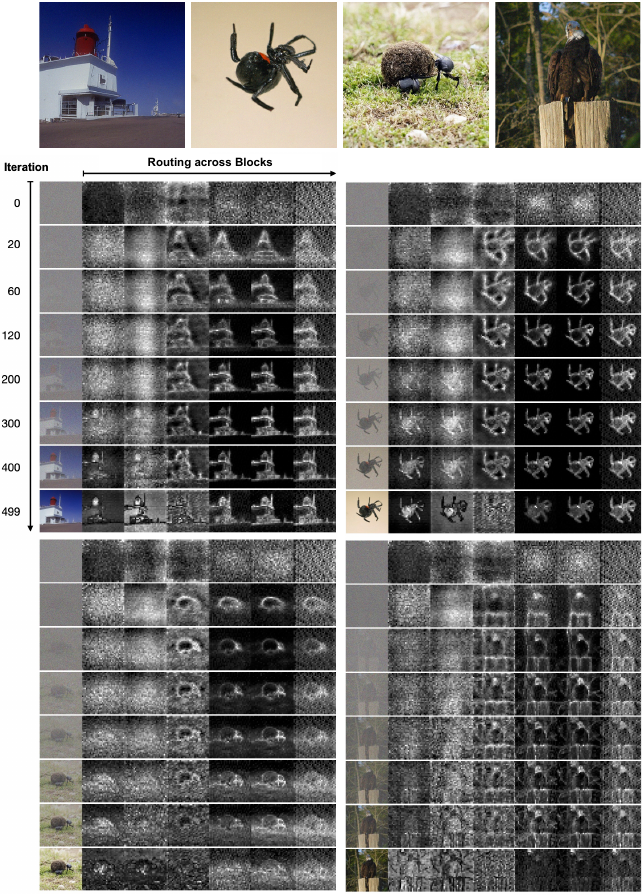}
\end{center}
\caption{\label{fig:in256_attn_1} Visualization of emergent adaptive computation for ImageNet $256\by 256$ samples.
}
\end{figure}
\begin{figure}[h!]
\begin{center}  
\includegraphics[width=0.75\linewidth]{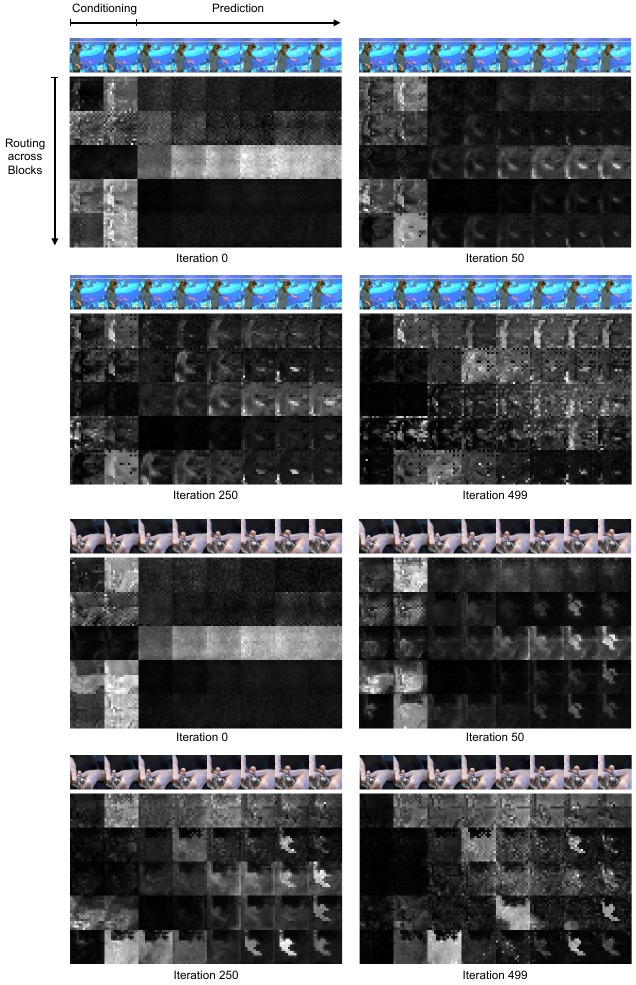}
\end{center}
\caption{\label{fig:kin_attn_1} Visualization of emergent adaptive computation for video prediction on Kinetics-600. The samples are subsampled 2$\times$ in time to align with the attention visualization. In each column of the attention visualization, the first two columns are read attention on conditioning frames. We observe that read attention and hence computation is focused on regions of motion, that cannot be generated by simply copying from the conditioning frames.
}
\end{figure}
\newpage

\end{document}